\documentclass[a4paper,fleqn]{cas-sc}

\usepackage[authoryear]{natbib}

\usepackage[most]{tcolorbox}
\usepackage{graphicx}
\usepackage{subcaption}
\usepackage{booktabs}
\usepackage{multirow}
\usepackage{hyperref}
\usepackage{fvextra}
\usepackage{tabularx}

\newtcolorbox{rqbox}{
  enhanced,
  colback=gray!18,
  colframe=black,
  boxrule=1.2pt,
  arc=3mm,
  left=4mm,
  right=4mm,
  top=3mm,
  bottom=3mm,
  before skip=8pt,
  after skip=8pt
}

\DefineVerbatimEnvironment{PromptBlock}{Verbatim}{
    breaklines=true,
    breakanywhere=true,
    fontsize=\small
}

\def\tsc#1{\csdef{#1}{\textsc{\lowercase{#1}}\xspace}}
\tsc{WGM}
\tsc{QE}

\begin{document}
\let\WriteBookmarks\relax
\def\floatpagepagefraction{1}
\def\textpagefraction{.001}

\shorttitle{}    

\shortauthors{Yang et al.}

\title [mode = title]{Beyond Clumsy Agentic Tool Calling: How Cueing and Spotting Make the Difference}  

\author[1]{Xindi Yang}
\ead{2534132@tongji.edu.cn}
\credit{Conceptualization, Methodology, Software, Investigation, Formal analysis, Writing - original draft}

\author[1]{Yan Liu}
\cormark[1]
\ead{yanliu.sse@tongji.edu.cn}
\cortext[1]{Corresponding author}
\credit{Supervision, Methodology, Writing - review and editing}

\author[2]{Jiajun Sun}
\ead{sunjj@shec.com.cn}
\credit{Supervision, Writing - review and editing}

\affiliation[1]{
  organization={School of Computer Science and Technology, Tongji University},
  city={Shanghai},
  country={China}
}

\affiliation[2]{
  organization={Shanghai Highly Electrical Appliance Co., Ltd},
  city={Shanghai},
  country={China}
}


\begin{abstract}
LLM-based task-oriented agents depend on tools to interact with their environments, but may overlook execution feedback, repeat ineffective calls, or persist with unproductive strategies. These behaviors can hinder agents' ability to use tools autonomously and sustain task progress. We characterize tool-use symptoms at two temporal granularities: local \emph{clumsy tool-use behaviors} and \emph{tool-use detours} spanning multiple interactions. Towards greater autonomy in tool use, we present ToolPulse, an online middleware supporting autonomous tool use through cueing and spotting. ToolPulse tracks execution state and temporal interaction patterns, monitoring proposed calls before dispatch to tools and inspecting execution feedback prior to the agent resumes reasoning. It combines state-based rules, semantic sequence analysis, and interaction-feature similarity to identify intervention opportunities and provide verbal reminders, structured feedback, or context-specific advice from auxiliary models. Our process-oriented evaluation distinguishes trigger validity, behavioral cessation, and substantive correction alongside task outcomes and resource costs. Using SWE-agent with three backbone models, we evaluate 1,350 trajectories on 90 Python, C++, and Java bug-fix tasks, comparing unassisted execution with four assistance strategies. Substantive responses follow 61.6--90.9\% of valid triggers for tool-use detours. Concise reminders achieve the highest observed aggregate substantive-response rate; more elaborate assistance offers symptom-dependent benefits and can be less effective. Aggregate repair outcomes remain close to baseline, with token savings in selected settings. These findings support proactive runtime assistance for autonomous tool use. The observed gap between behavioral adjustment and aggregate repair gains calls for careful decisions on intervention timing and assistance strategy in autonomous agents.
\end{abstract}




\maketitle

\section{Introduction}

In recent years, LLM-based task-oriented agents have emerged as autonomous actors that perceive, reason about, and interact with their environments to solve problems. Tool use underpins the autonomy of task-oriented agents: tools provide the means through which agents acquire information, turning reasoning into task execution, act on external environments, and observe the consequences of their actions. These capabilities support applications in areas such as scientific research, data analysis, embodied interaction, and software engineering \citep{Guo2024DSAgent,Chen2026XRayScientist,Sahu2025InsightBench,Yang2024SWEAgent,Chen2025EMOS}. Task-oriented agents typically rely on tool documentation in their context, including functional descriptions, parameter schemas, and a few invocation examples, to understand what tools do and how to use them \citep{Wang2025ToolGen,Xu2025ToolLearningSurvey,Yuan2025EasyTool}. However, knowing a tool’s name, interface, or input format does not ensure that the agent can use the tool autonomously as a task unfolds. Tool descriptions convey basic declarative knowledge but cannot cover every situation encountered during execution. Moreover, agents often treat tool output as transient context for the next reasoning step in a loop of execution, observation, and further execution. The returned output is appended to the interaction history as an observation or tool-response message, leaving the model to interpret the feedback \citep{Yang2024SWEAgent,wang2026openhandssoftwareagentsdk}. When the agent fails to relate this feedback to the evolving execution state and earlier interactions, it may miss diagnostic signals, invoke an unnecessarily costly tool, or persist with a strategy that has ceased to work.

Problematic tool-use behaviors are difficult to capture and understand from individual tool calls or isolated outputs, as they depend on the execution context and how interactions unfold over time. Tool use is commonly understood in terms of call accuracy and task success \citep{li-etal-2023-api,pmlr-v267-patil25a,yao2024taubenchbenchmarktoolagentuserinteraction}, while how problematic behaviors emerge and persist during execution remains less clearly characterized. We identify two types of symptoms of difficulties in autonomous tool use: \emph{clumsy tool-use behaviors} and \emph{tool-use detours}. Clumsy tool-use behaviors are local interactions that are unnecessary or poorly matched to the available evidence and execution context. They include overlooking diagnostic information in tool feedback, invoking an unnecessarily costly tool, or repeating a failed call without addressing its reported cause. Tool-use detours occur across a sequence of interactions, when an agent persists with an unproductive strategy despite feedback that calls for adjustment. Such detours may involve recurring action patterns or repeated errors that the agent fails to use to revise its approach. Local missteps can accumulate into longer detours, while individually valid calls can also form an unproductive sequence. Both symptoms can undermine the agent's ability to adjust its tool use and carry a task forward autonomously, potentially increasing execution costs. Mitigating these symptoms calls for adaptive support that responds to the agent’s evolving execution context and interaction patterns, helping it recognize when and how to adjust its tool use.

Recent work has explored trajectory refinement, experience reuse, and offline training as approaches to improving agent capabilities \citep{antoniades_swe-search_2025,mu_experepair_2026,Pan2025SWEGym}. Their benefits depend on additional exploration, applicable historical experience, or parameter updates, motivating complementary support during execution. More directly, Wink periodically inspects agent trajectories and injects corrective guidance \citep{Nanda2026Wink}, but emerging problems may remain unaddressed between inspections. LivePlan uses predefined rules to trigger corrective advice \citep{liu2026onlinemonitoringcorrectivesteering}. However, its reliance on predefined rules can make it difficult to detect stagnation manifested through approximate repetition across recent interactions. These approaches demonstrate the value of online intervention while motivating closer attention to how tool interactions develop within the evolving execution state. We therefore focus on combining local evidence about tool use with temporal patterns across interactions to recognize clumsy tool-use behaviors and tool-use detours, and provide targeted assistance at the relevant interaction points.

Although current task-oriented agents vary widely in architecture and workflow, their tool use typically involves a common cycle of invoking tools, interpreting feedback, and selecting subsequent tools. Investigating how to strengthen agents' ability to use tools autonomously therefore requires concrete tasks through which their tool interactions and responses to runtime assistance can be observed and evaluated. Software engineering provides a representative setting for studying these symptoms because coding agents depend on sustained tool use to inspect and modify a repository while responding to execution feedback. Agents such as SWE-agent \citep{Yang2024SWEAgent} and AutoCodeRover \citep{zhang_autocoderover_2024} use tools to browse and edit code, compile and test programs, and manage versions. To fix a bug, for instance, an agent must locate relevant code, gather diagnostic information, modify the program, and validate the patch. Throughout this process, the suitability of a tool call depends on what has already been inspected, what has changed, and what earlier executions have returned. These observable changes in repository state together with test outcomes, make software engineering particularly suitable for examining how online assistance can help agents adjust their tool use and carry tasks forward autonomously. We therefore investigate how runtime assistance helps agents adjust their tool use and sustain autonomous task progress, using SWE-agent on repository-level bug-fix tasks as the experimental environment.

\begin{figure}
	\centering
	\includegraphics[
        width=\linewidth,
        height=0.70\textheight,
        keepaspectratio
    ]{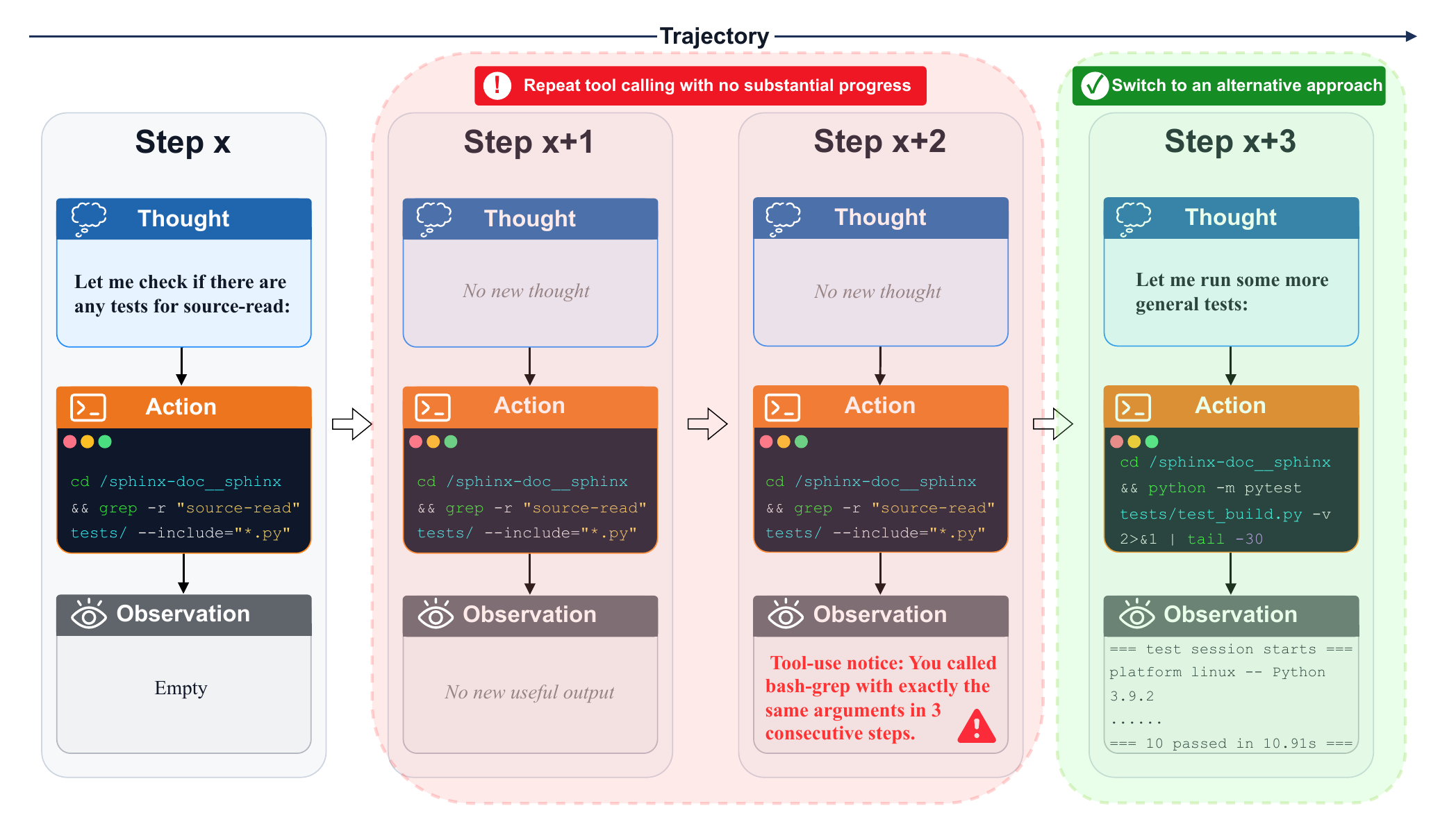}
    \caption{An example of ToolPulse's verbal cueing, showing repeated search calls, a runtime reminder, and the agent's subsequent switch to testing.}
    \label{fig:traj}
\end{figure}

Natural language offers a way to support these adjustments by drawing agents’ attention to relevant execution evidence and guiding their subsequent tool-use decisions. To explore how online linguistic intervention can support tool use while preserving agents' decision-making autonomy, we introduce ToolPulse, a state-flow-tracking online tool-use assistance middleware designed as a fundamental support layer for autonomous tool use in task-oriented agents. ToolPulse tracks the agent's execution context and temporal tool-use patterns across interactions. For each interaction, it monitors proposed calls before dispatch to tools and inspects execution feedback prior to the agent resumes reasoning. ToolPulse uses the evolving state, which allows it to adapt to the agent's working rhythm, to identify clumsy tool-use behaviors and tool-use detours. When intervention is warranted, ToolPulse can intercept an insufficiently grounded call and deliver targeted assistance to the agent. Instead of imposing rigid, mechanical corrections or simply accumulating context, ToolPulse offers adaptive assistance in three forms: \emph{verbal cueing} through brief reminders or guidance, \emph{visual cueing} through structured presentation of textual information, and \emph{tactile spotting} through context-specific suggestions from a fast auxiliary language model. As a weightlifting spotter would, it offers a cue, or lends a hand when needed while the agent remains responsible for carrying the task forward. Figure~\ref{fig:traj} illustrates ToolPulse's verbal cueing during a bug-fix task. After the agent issues the same search command in three consecutive steps, ToolPulse adds a reminder to the observation, and the agent subsequently switches to running tests.

The main contributions of this work are as follows:
\begin{itemize}
    \item We characterize difficulties in autonomous tool use at two temporal granularities: local \emph{clumsy tool-use behaviors} and \emph{tool-use detours} spanning multiple interactions. Guided by a pilot study, we organize six symptoms along exploration, progress, grounding, and recovery, providing operational criteria for online monitoring and offline trajectory assessment.

    \item We introduce ToolPulse, an online middleware that detects tool-use symptoms through state-flow tracking and provides four assistance strategies: reminder-only verbal cueing, guided verbal cueing, visual cueing, and tactile spotting. It combines state-based rules, semantic sequence analysis, and interaction-feature similarity to provide selective, minimally intrusive assistance that helps agents adapt their tool use as tasks unfold.

    \item We empirically investigate ToolPulse across 1,350 trajectories on 90 bug-fix tasks in Python, C++, and Java using three backbone models. Comparing different assistance strategies, we document substantive behavioral adjustments following assistance and examine how different forms of support affect agents' responses to tool-use difficulties. The findings indicate that concise reminders can be sufficient to elicit substantive adjustment, while more elaborate assistance offers symptom-dependent benefits and can even be less effective than concise reminders. We additionally assess task outcomes and resource costs to examine the practical effects of runtime support.

    \item We propose a reference implementation of ToolPulse and document the task instances, detection settings, assistance prompts, and assessment procedures to support reproduction and further investigation of runtime support for autonomous tool use.
\end{itemize}

We begin with a pilot study to explore the following question: \textbf{RQ0)} \textit{Can additional natural-language guidance help task-oriented agents improve their subsequent tool-use behavior?} Informed by the pilot findings, we design ToolPulse and conduct experiments to evaluate its effects. The remainder of this paper is organized as follows. After reviewing related work, Section~\ref{sec3} presents the pilot study and its implications for ToolPulse. Section~\ref{sec4} characterizes the tool-use symptoms and introduces ToolPulse's monitoring and assistance mechanisms. Section~\ref{sec5} evaluates behavioral correction, compares assistance forms, and examines task outcomes and resource costs. The subsequent sections discuss ToolPulse's role in agent harness, examine threats to validity, and conclude the paper.

\section{Related work}\label{sec2}
\subsection{Tool Use and Interaction Interfaces}

The tool-use capabilities of LLM-based agents depend not only on how the backbone model selects and composes actions, but also on the design of the action space, interaction interface, and presentation of environmental feedback. CodeAct uses executable Python code to consolidate an agent's actions into a unified action space, enabling the agent to compose multiple tools and revise or generate actions in response to new observations \citep{Wang2024CodeAct}. SWE-agent introduces a custom agent-computer interface (ACI) that enables agents to navigate repositories, edit code files, and execute tests and other programs, and investigates how interface design affects agent behavior and performance \citep{Yang2024SWEAgent}. AutoCodeRover performs structure-aware code search over an abstract-syntax-tree-based program representation and, when a test suite is available, uses spectrum-based fault localization to further narrow the context relevant to the root cause \citep{zhang_autocoderover_2024}. In web tasks, AgentOccam further shows that refining the observation and action spaces to better align with the capabilities of the backbone model can improve task performance \citep{Yang2025AgentOccam}.

Another line of research improves agents' ability to perform tasks in tool-mediated environments through training. SWE-Gym introduces an executable environment for training real-world software engineering agents and trains verifiers on agent trajectories to support inference-time scaling \citep{Pan2025SWEGym}. R2E-Gym studies the scalable, procedural curation of executable training environments and combines execution-based and execution-free verifiers for hybrid test-time scaling \citep{Jain2025R2EGym}.

The studies reviewed above improve agents' tool use through interaction design and training in executable environments. ToolPulse complements these approaches with runtime support for autonomous tool use. It tracks execution state and interaction patterns to identify tool-use difficulties as they emerge, then provides targeted assistance to help agents interpret feedback and adjust subsequent actions.

\subsection{Reflection and Experience Reuse}

Reflection mechanisms revisit an agent's actions and the resulting feedback to inform subsequent decisions. Devil's Advocate introduces anticipatory reflection before action execution, post-action alignment with subtask objectives, and a comprehensive review upon plan completion to anticipate potential failures, assess the suitability of executed actions, and refine future strategies \citep{Wang2024DevilsAdvocate}. Such methods show that feedback-driven adjustment can be incorporated at multiple points during task execution.

Memory-augmented methods further investigate how experience from past interactions can be retained and reused. Agent Workflow Memory induces reusable routines, termed workflows, from interaction trajectories and supports both offline induction from training examples and online induction from test queries on the fly \citep{Wang2025AgentWorkflowMemory}. Dynamic Cheatsheet maintains a persistent, evolving external memory at inference time, storing accumulated strategies, code snippets, and general problem-solving insights for reuse across queries \citep{Suzgun2026DynamicCheatsheet}. The former primarily captures reusable action procedures, whereas the latter focuses on accumulating and reusing concise problem-solving knowledge across tasks.

Subsequent studies further examine how experience should be represented, organized, and updated. Agentic Context Engineering treats context as an evolving playbook that accumulates, refines, and organizes strategies through generation, reflection, and curation, thereby supporting both offline and online adaptation \citep{Zhang2026ACE}. Experiential Reflective Learning reflects on task trajectories and outcomes to generate transferable heuristics, and retrieves heuristics relevant to the current task at test time to guide execution \citep{Allard2026ERL}. Hu et al. show that moving continual learning to external memory does not eliminate the challenges of experience reuse. Under a limited context window, old and new experiences compete during retrieval, while memory representation and retrieval organization jointly affect transfer and forgetting \citep{Hu2026ContinualLearningMemory}.

Reflection and experience reuse provide agents with feedback-derived knowledge for subsequent decisions. ToolPulse shares this feedback-driven perspective and focuses on assistance triggered by observable tool-use symptoms within the ongoing task. It relates proposed actions and execution feedback to the accumulated interaction state, using this evidence to help agents recognize difficulties and revise their tool use at the relevant decision points.

\subsection{Online Feedback and Execution Assistance}

Execution-trajectory analysis provides a basis for identifying behavioral problems and determining when an agent may require assistance. Bouzenia and Pradel conduct a large-scale empirical study of software engineering agents' thought-action-result trajectories, analyzing trajectory structure, action patterns, and token usage, together with reasoning coherence and feedback integration \citep{Bouzenia2025SEAgentTrajectories}. AgentRx further automates the diagnosis of failed trajectories by synthesizing constraints, evaluating them step by step, and producing an auditable validation log of constraint violations, which is then used to localize the critical failure step and its category \citep{Barke2026AgentRx}. The former focuses on characterizing agent behavior and anti-patterns, whereas the latter focuses on attributing failures in trajectories that have already failed. Wink asynchronously inspects agent trajectories at fixed step intervals to detect Specification Drift, Reasoning Problems, and Tool Call Failures, and injects targeted course-correction guidance into the coding agent; its production-scale evaluation provides empirical support for runtime behavioral assistance and a broader-grained precedent for our distinction between local clumsy tool-use behaviors and persistent tool-use detours \citep{Nanda2026Wink}. LivePlan uses predefined rules to monitor execution patterns and trigger corrective advice during task execution \citep{liu2026onlinemonitoringcorrectivesteering}.

Other work grounds recovery in runtime evidence associated with tool-use and execution failures. PROBE organizes telemetry from failed runs into structured evidence, structured diagnosis, and bounded recovery guidance for a subsequent attempt \citep{Zhao2026PROBE}. Outcome Monitors target silent tool failures in which an output arrives in the expected format but violates an outcome contract. They preserve the original result and issue a nonbinding receipt that identifies the violated property and the available public recovery tools, leaving the executing agent to determine the next action \citep{Panthi2026OutcomeMonitors}. Dubey trains lightweight anomaly monitors on step-level telemetry and combines them with deterministic verification, rollback, and live re-execution to connect failure detection with repair \citep{Dubey2026AgentFailureRepair}.

These studies establish execution evidence as a basis for behavioral diagnosis and runtime assistance. ToolPulse combines local execution state with temporal interaction patterns to address clumsy tool-use behaviors and persistent tool-use detours. Its pre- and post-execution checkpoints connect proposed calls and returned feedback with targeted cueing and spotting. Under shared detection settings, we examine how four assistance strategies affect agents' substantive tool-use adjustments, investigating how runtime support can help agents sustain autonomous task progress.

\section{Learning by Doing: a Half-successful Story}\label{sec3}

To identify effective mechanism of online linguistic intervention that help agents adjust their tool use autonomously, we first conduct a pilot study to investigate a more basic question: whether a coding agent's tool-use behavior can be improved through additional natural-language interventions during task execution. It is natural to consider experience-based interventions that help agents reflect on their tool interactions and adjust subsequent actions. This idea is grounded in experiential learning theory. Human engineers often become proficient with tools through sustained practice. Dewey’s notion of learning by doing \citep{dewey1916democracy} emphasizes acquiring knowledge through action; building on this tradition, Kolb formulated an experiential learning cycle \citep{kolb1984experiential} comprising Concrete Experience, Reflective Observation, Abstract Conceptualization, and Active Experimentation. This cycle offers a useful lens for learning to use software tools. Engineers act and observe the result; when problems arise, they revisit their actions, locate errors, consult documentation, and turn their observations and new knowledge into principles that they can test and revise in subsequent work.

At this stage, natural language still remains a common medium for guiding agents. To study how linguistic guidance affects tool use during an ongoing task, we adopt the established ReAct interaction pattern \citep{yao_react_2023}. ReAct interleaves reasoning, actions, and observations in a recurring loop: at each step, the agent generates a thought and a command, executes the command, and incorporates the resulting feedback into subsequent decisions. As the agent carries forward the task through successive iterations of this loop, we take each reasoning--action--observation cycle as the basic unit of analysis and represent an individual task run as 
\begin{equation}
\tau = \bigl((r_t, a_t, o_t)\bigr)_{t=1}^{T}
\end{equation}
where $r_t$, $a_t$, and $o_t$ denote the thought, tool action, and observation at step $t$, respectively, and $T$ is the length of the run. This structure allows us to observe and assist tool use after feedback is returned and before the agent makes its next decision.

We devise an online experiential tool-learning component for task-oriented agents. It treats each tool interaction as an experience that can be examined. Using the agent’s tool call, its execution result, and the current task trajectory, the component determines whether the latest interaction warrants attention: whether it involves repetitive tool use or forms part of an unproductive sequence of actions. Upon detecting such a pattern, the component fills a predefined prompt template with selected details of the action and appends the resulting message to the current step’s observation. Before producing its next thought, the agent therefore sees both the outcome of its previous action and a prompt about that action. It can then reflect on the interaction, extract a lesson, and adjust subsequent actions within the ongoing task. The pilot study tests whether such immediate linguistic feedback can help an agent revise its tool use and aims to provide an empirical basis for design of effective online linguistic interventions that support autonomous tool use.

Although prior work \citep{Nanda2026Wink} has begun to examine misbehaviors in coding agents, existing studies and classifications remain broad. We still lack a fine-grained understanding and systematic classification of misbehaviors in coding agents' tool use. To tie online assistance to observable tool interactions, we draw on preliminary observations of baseline trajectories and expert judgment to identify four action patterns. We try these patterns in the pilot study to flag interactions that may warrant attention, while gaining an initial sense of whether they can serve as actionable triggers for subsequent online assistance. Duplicate Action occurs when the agent issues two consecutive tool calls with exactly the same tool and arguments. Loop Action occurs when the agent repeatedly executes the same or a similar ordered sequence of tools. Failed Action occurs when a tool call returns an execution error. Custom Tool Candidate denotes a reusable operation sequence that can be encapsulated as a custom tool.

The first three patterns mark tool interactions that may warrant reconsideration. Custom Tool Candidate instead identifies a positive opportunity to improve reuse through tool abstraction; it does not indicate anomalous behavior. To support this process, we provide the agent with two auxiliary tools: one creates custom tools, and the other invokes previously created custom tools. With this setup, we task the agent with fixing software bugs and examine how the interventions affect its tool use during task execution.

\subsection{Study Setup}

\paragraph{Agent and Models.} 
We use SWE-agent as the base agent framework. Through an explicit Agent-Computer Interface, SWE-agent provides the language model with operations for code search, file navigation, program editing, and test execution. After each action, the environment returns feedback, yielding a structured record of tool interactions \citep{Yang2024SWEAgent}.

We use DeepSeek-V4-Pro \citep{deepseekai2026deepseekv4} and Qwen3-Coder-Plus \citep{qwen3technicalreport} as the language models underlying SWE-agent. DeepSeek-V4-Pro is a general-purpose model with capabilities in reasoning, code generation, and agentic tool use, whereas Qwen3-Coder-Plus is optimized for code generation and coding-agent tasks, with an emphasis on multistep interaction with the environment and tool use. Their different design emphases allow us to examine, on a preliminary basis, whether the effect of linguistic intervention is confined to one type of backbone model. All experiments use the same SWE-agent architecture and tool environment.

\paragraph{Task.} 
We draw our experimental tasks from the bug-fix subset of OmniCode. OmniCode is a multitask benchmark for software engineering agents that covers Python, Java, and C++ and includes bug-fix, test generation, code review fixes, and style fixes \citep{sonwane_omnicode_2026}. These tasks make agents' tool-use behavior observable through the tool choices and action sequences they employ to accomplish software engineering objectives. A bug-fix task requires the agent to understand an issue in an executable repository environment, locate relevant code, produce a patch, and validate the changes through testing. The successive activities of retrieval, editing, and validation make these tasks suitable for observing tool use during long-horizon task runs.

Given the exploratory aim and computational cost, we select 30 instances from the Python bug-fix subset. To sample tasks across a range of complexity, we construct a proxy score from each instance's gold patch, test patch, and issue description. Specifically, we extract the number of added and deleted lines in the gold patch (\textit{patch\_churn}), the number of added and deleted lines in the test patch (\textit{test\_churn}), the number of files changed by the gold patch (\textit{patch\_files}), and the character count of the issue description (\textit{problem\_chars}). We convert each feature to a percentile rank in $[0,1]$ and sum the four ranks:
\begin{equation}
C_i =
R_i^{\mathrm{patch\_churn}}
+ R_i^{\mathrm{test\_churn}}
+ R_i^{\mathrm{patch\_files}}
+ R_i^{\mathrm{problem\_chars}}
\end{equation}
where $R_i^x$ denotes the percentile rank of instance $i$ for feature $x$. We then divide the tasks into easy, medium, and hard level according to their composite scores and randomly select 10 instances from each stratum. This procedure reduces the chance that the pilot sample is concentrated in one complexity range. The score is used only for sampling; we do not treat it as a direct measure of actual task difficulty.

\paragraph{Comparison Settings.}
For each backbone model, we compare a baseline with three intervention groups. The baseline runs the original SWE-agent without additional linguistic guidance. The intervention groups use the same prompting procedure but differ in their trajectory representations and sequence detection methods. Each model is evaluated under all four configurations on the same 30 tasks, with one run per model–task–configuration combination. Table~\ref{tab:pilot_groups} summarizes the trajectory representations and sequence detection methods used by the three intervention groups.

\begin{table}[t]
\caption{Trajectory representations and sequence detection methods in the pilot study.}
\label{tab:pilot_groups}
\centering
\small
\begin{tabularx}{\linewidth}{@{}lXX@{}}
\toprule
\textbf{Group} & \textbf{Trajectory Representation} & \textbf{Sequence Detection} \\
\midrule
G1 &
A sequence of discrete symbols, each representing a tool name. &
Sequitur identifies recurring and nested tool-name sequences through incremental hierarchical grammar induction \citep{nevill-manning_identifying_1997}. \\
\addlinespace
G2 &
A sequence of bug-fix-oriented labels assigned to interactions according to their function and execution result. &
Sequitur identifies recurring functional patterns, allowing calls with different tool names or arguments to share the same symbolic representation. \\
\addlinespace
G3 &
A sequence of numerical feature vectors describing model behavior and task state. &
Matrix Profile identifies approximately recurring trajectory segments through nearest-neighbor distances between fixed-length subsequences \citep{yeh_matrix_2016}. \\
\bottomrule
\end{tabularx}
\end{table}

For G1 and G2, Sequitur's incremental operation accommodates the stream of interactions generated during task execution. For G2, we organize tool use around three purposes: Retrieve obtains task-relevant information from the environment; Compute calculates or validates results; and Operate changes the environment state. Inspired by prior work on encoding coding-agent trajectories \citep{Bouzenia2025SEAgentTrajectories,mehtiyev_beyond_2026}, we define fine-grained labels for bug-fix tasks and assign each interaction a single label using deterministic rules based on the invoked tool and its execution result. Table~\ref{tab:bugfix_labels} presents these labels and their definitions.

For G3, model-behavior features include thought length, tool identity, invocation arguments, and argument length. Task-state features include observation length, tool execution time, and error presence. Length and time features undergo logarithmic transformation and online standardization, while tool names and arguments use fixed-dimensional multi-hash encodings. These features support comparison of trajectory segments even when their tool names differ.

\begin{table}[t]
\caption{Bugfix-oriented action labels and their definitions.}
\label{tab:bugfix_labels}
\centering
\begin{tabular*}{\linewidth}{@{\extracolsep{\fill}}lll@{}}
\toprule
\textbf{Functional Level} & \textbf{Label} & \textbf{Meaning} \\
\midrule
Retrieve & Ls  & search \\
         & Lt  & targeted read \\
         & Lb  & browse whole file \\
\addlinespace
Operate  & E   & environment setup \\
         & P   & edit \\
         & Ps  & edit with syntax error \\
         & Pi  & edit with import error \\
         & Pr  & re-edit file \\
\addlinespace
Compute  & Vp  & test run with no error \\
         & Vf  & test run with failure \\
         & Ve  & test run with error \\
\addlinespace
Other    & G   & general / other \\
         & Emp & empty action \\
\bottomrule
\end{tabular*}
\end{table}

When a detector finds a target pattern, the component extracts relevant information from recent actions and their execution results, inserts it into the corresponding prompt template, and appends the prompt after the observation. The prompt asks the agent to examine the tool interaction that just occurred, assess whether its current course still advances the task, and revise its next step if needed. The intervention does not decide on the agent’s behalf: the agent continues to generate its thoughts, select tools, and execute actions.

\paragraph{Evaluation Measures.} We analyze task outcomes and trigger details in the pilot study. 

For task outcomes, we count the tasks for which the agent successfully submits a patch, the tasks for which the patch passes the tests, and token usage. Token usage is the sum of the backbone model's input and output tokens over the complete trajectory. For intervention strategy $s$ and its set of instances $\mathcal I$ paired with the baseline, the relative change in token usage is calculated as follows, 
\begin{equation}\label{eq:pilot_token_change}
\Delta_T(\mathcal I,s)\,[\%] = 100\, \frac{\sum_{i\in\mathcal I}T_{s,i} -\sum_{i\in\mathcal I}T_{\mathrm{baseline},i}} {\sum_{i\in\mathcal I}T_{\mathrm{baseline},i}}
\end{equation}
where $T_{s,i}$ is the token usage recorded for strategy $s$ on instance $i$. A negative value indicates lower usage than the paired baseline.

For trigger details, we record how often each of the four action is captured and assess detection quality. Depending on the pattern, quality is checked through scripted validation or review of sampled trajectories. We treat these findings as exploratory evidence of feasibility.

The complete prompt templates, task list, and run configurations for the pilot study are provided in Appendix~\ref{secA1}.

\subsection{Result}
We first evaluate how runtime natural-language interventions affect task completion.  We observe that the agent successfully submitted a patch in all tasks, none of the three intervention strategies impairs the agent's basic ability to complete the task. Then we measure task outcome by whether the submitted patch passes the issue-specific tests. As shown in Table~\ref{tab:pilot_task_outcomes}, the differences between the intervention groups and the baseline are small in terms of pass rate. For DeepSeek-V4-Pro, the pass rates range from 70.0\% to 73.3\% across the four groups. For Qwen3-Coder-Plus, they range from 56.7\% to 63.3\%. Although the interventions change the outcomes of some tasks, enabling the agent to solve some tasks that remain unresolved under the baseline while causing a small number of previously solved tasks to fail, the improvements and regressions offset each other to some extent. These small differences are insufficient to show that any intervention consistently improves the task resolution. \textbf{Finding 1. Runtime natural-language interventions preserve the agent's ability to complete tasks, but the uncertainty in their effects on final task resolution deserves attention.}

\begin{table}[t]
\caption{Performance comparison under different intervention groups. \textit{Task} is the number of bug-fix instances evaluated for each LLM. \textit{Baseline} is the number of instances whose patch passes the issue-specific tests without intervention. Each intervention strategy cell has the form $P\,(G/L)$: $P$ is the number passing under that strategy, $G$ is the number that fails under baseline but passes with intervention (newly passed), and $L$ is the number that passes under baseline but fails with intervention (regressed). All comparisons use the same task instances within each LLM.}
\label{tab:pilot_task_outcomes}
\centering
\small
\resizebox{\linewidth}{!}{%
\begin{tabular}{lccccc}
\toprule
\textbf{LLM}
& \textbf{Task}
& \textbf{Baseline}
& \multicolumn{3}{c}{\textbf{Intervention Strategy}} \\
\cmidrule(lr){4-6}
&
&
&
\textbf{Sequitur}
& \textbf{Semantic Sequitur}
& \textbf{Matrix Profile} \\
\midrule

DeepSeek-V4-Pro
& 30
& 21
& 21 (2/2)
& 22 (1/0)
& 21 (2/2) \\

Qwen3-Coder-Plus
& 30
& 19
& 18 (1/2)
& 18 (0/1)
& 17 (0/2) \\

\bottomrule
\end{tabular}%
}
\end{table}

\begin{figure}[pos=t]
    \centering

    \begin{subfigure}{0.75\linewidth}
        \centering
        \includegraphics[width=\linewidth]{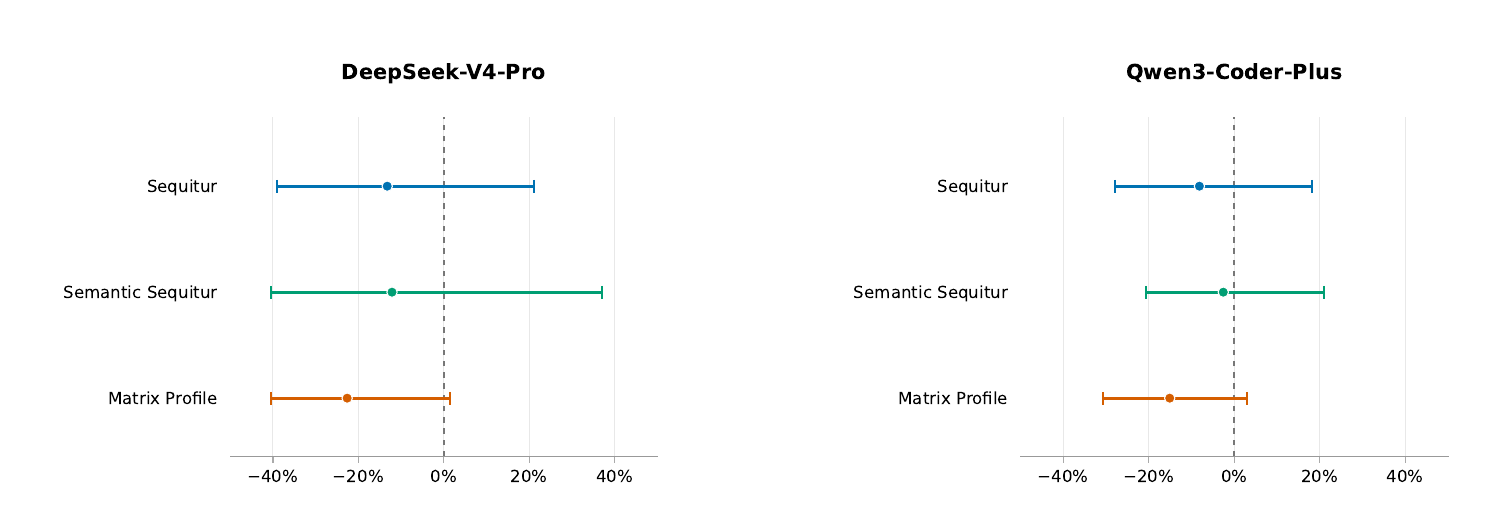}
        \caption{Overall relative change in total token usage across intervention strategies and backbone models.}
        \label{fig:method_a}
    \end{subfigure}

    \par\vspace{0.5em}

    \begin{subfigure}{0.75\linewidth}
        \centering
        \includegraphics[width=\linewidth]{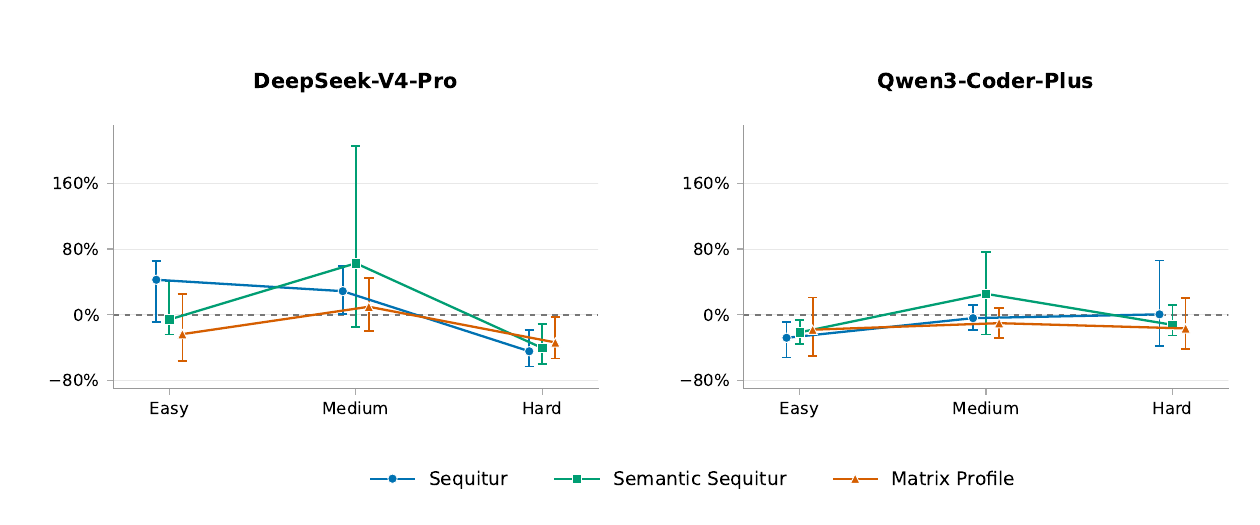}
        \caption{Relative change in total token usage across task difficulty levels, intervention strategies, and backbone models.}
        \label{fig:method_b}
    \end{subfigure}

    \caption{Relative changes in total token usaage against the paired baseline for different intervention strategies under two backbone LLMs. Points show relative changes in summed tokens, negative values indicate savings, and error bars show 95\% paired bootstrap confidence intervals obtained by resampling instances with replacement, retaining baseline--method pairs, and recomputing the relative change; resampling in (b) is conducted within each difficulty stratum.}
    \label{fig:pilot_comparison}
\end{figure}

We next investigate whether natural-language interventions reduce the cost of token, using the relative change $\Delta_T$ defined in Eq.~\ref{eq:pilot_token_change}. Fig~\ref{fig:pilot_comparison}(a) shows negative overall point estimates for all three methods on both backbones, with Matrix Profile showing the largest estimated reduction. However, all 95\% paired bootstrap confidence intervals include zero and extend into positive values, so the data remain compatible with both no change and increased token use. Fig~\ref{fig:pilot_comparison}(b) shows stronger evidence of savings on hard tasks with DeepSeek-V4-Pro, where all three intervals lie below zero, while results on easy and medium tasks are mixed. For Qwen3-Coder-Plus, the hard-task estimates are near zero or moderately negative, but their intervals include zero. Evidence of token savings is thus strongest on hard tasks with DeepSeek-V4-Pro and varies with the backbone and task difficulty. The results show that runtime natural-language interventions do not consistently improve pass rates, but produce observable cost benefits under several experimental conditions, suggesting that natural-language prompts can alter an agent's subsequent actions under some conditions and reduce the interaction overhead required to complete a task. \textbf{Finding 2. Runtime natural-language interventions provide preliminary evidence of supporting agents in adjusting their tool use during task execution, with reduced interaction costs observed under several experimental conditions.}

To examine whether natural-language interventions actually change how agents use tools, we analyze all trajectories offline, replaying each detector on baseline runs without prompts and rescanning intervention runs with repeated-alert suppression disabled. Comparisons are restricted to the same detector and task set because Loop Action definitions differ across methods.

\begin{table}[pos=t]
\caption{\textit{Escape} is the percentage of Duplicate Action triggers followed by a different tool call at the next step.
\textit{Recurrence} is the percentage of task instances with Duplicate Action in which it recurs later.
\textit{Any-pattern} and \textit{Same-pattern} are the percentages of Loop Action triggers followed by another Loop Action of any pattern or the same pattern, respectively, in the remaining trajectory.
The three baseline entries in each Loop Action column correspond to Sequitur, Semantic Sequitur, and Matrix Profile, from top to bottom.
\textit{Mean} is the mean Failed Action count per task instance containing at least one Failed Action; \textit{Max} is the largest within-instance count of any single tool--error-type pair.
Parentheses give numerator/denominator counts in the corresponding statistical unit; n/a indicates no eligible triggers or task instances.}
\label{tab:pilot_behaviors}
\centering
\small
\setlength{\tabcolsep}{3pt}
\renewcommand{\arraystretch}{1.15}

\resizebox{\linewidth}{!}{%
\begin{tabular}{llcccccc}
\toprule
\multirow{2}{*}{\textbf{Model}}
& \multirow{2}{*}{\textbf{Group}}
& \multicolumn{2}{c}{\textbf{Duplicate Action (\%)}}
& \multicolumn{2}{c}{\textbf{Loop Action (\%)}}
& \multicolumn{2}{c}{\textbf{Failed Action}} \\
\cmidrule(lr){3-4}
\cmidrule(lr){5-6}
\cmidrule(lr){7-8}
& 
& \textbf{Escape}
& \textbf{Recurrence}
& \textbf{Any-pattern}
& \textbf{Same-pattern}
& \textbf{Mean}
& \textbf{Max} \\
\midrule

\multirow{4}{*}{DeepSeek-V4-Pro}
& Baseline
& 100.00 (3/3)
& 0.00 (0/3)
& \shortstack{33.33 (1/3)\\0.00 (0/1)\\88.89 (32/36)}
& \shortstack{0.00 (0/3)\\0.00 (0/1)\\61.11 (22/36)}
& 8.66
& 22 \\

\addlinespace

& Sequitur
& 80.00 (4/5)
& 0.00 (0/5)
& 0.00 (0/2)
& 0.00 (0/2)
& 8.17
& 46 \\

& Semantic Sequitur
& 58.33 (7/12)
& 11.11 (1/9)
& 0.00 (0/3)
& 0.00 (0/3)
& 7.76
& 33 \\

& Matrix Profile
& 88.89 (8/9)
& 12.50 (1/8)
& 75.00 (21/28)
& 50.00 (14/28)
& 7.55
& 24 \\

\midrule

\multirow{4}{*}{Qwen3-Coder-Plus}
& Baseline
& N/A
& N/A
& \shortstack{25.00 (1/4)\\0.00 (0/1)\\84.21 (16/19)}
& \shortstack{0.00 (0/4)\\0.00 (0/1)\\63.16 (12/19)}
& 7.30
& 11 \\

\addlinespace

& Sequitur
& N/A
& N/A
& 0.00 (0/1)
& 0.00 (0/1)
& 6.90
& 22 \\

& Semantic Sequitur
& 100.00 (1/1)
& 0.00 (0/1)
& N/A
& N/A
& 7.52
& 8 \\

& Matrix Profile
& 50.00 (1/2)
& 0.00 (0/2)
& 76.47 (13/17)
& 35.29 (6/17)
& 6.66
& 11 \\

\bottomrule
\end{tabular}%
}
\end{table}

Early reminders show no clear additional benefit. All three occurrences of Duplicate Action in the baseline of DeepSeek-V4-Pro are followed by immediate escape and no later recurrence, whereas immediate escape rates range from 58.33\% to 88.89\% under intervention (Table~\ref{tab:pilot_behaviors}). Duplicate Action recurs under Semantic Sequitur and Matrix Profile. Qwen3-Coder-Plus exhibits no Duplicate Action under the baseline and at most two occurrences of Duplicate Action per intervention group. Given the small sample size and the independent generation of trajectories across groups, these results do not demonstrate that natural-language interventions impair recovery. They do, however, indicate that the agent exhibits some ability to recover autonomously from simple, explicit anomalies. They motivate delaying subsequent intervention from two to three consecutive calls with identical tool names and arguments, thereby allowing the agent an opportunity to correct local mistakes independently, which also supported by recent research \citep{duan2026interveneempiricalstudyerrors}.

Similarity-based monitoring shows the clearest descriptive contrast in subsequent detection of Loop Action. Sequitur and Semantic Sequitur capture few occurrences of Loop Action. Further examination indicates that encoding actions using only tool names removes critical semantics and is insufficient to distinguish reasonable iterative operations from unproductive loops. Under Matrix Profile, the baseline subsequent-redetection rates are 88.89\% for DeepSeek-V4-Pro and 84.21\% for Qwen3-Coder-Plus; the corresponding same-pattern rates are 61.11\% and 63.16\%. Both rates show a moderate decrease under intervention (Table~\ref{tab:pilot_behaviors}). These results suggest that natural-language interventions may encourage the agent to leave regions of sustained action loop.

We further conduct an examination of a randomly sampled pair of baseline and Matrix Profile trajectories for a task mwaskom-seaborn-3187. Of the three trigger points in the Matrix Profile trajectory, two are manually judged to be reasonable. After receiving the guidance, the agent shifts from writing numerous exploratory scripts to inspecting specific code and from repeatedly writing temporary scripts to validating existing modifications. Offline replay of the baseline trajectory identifies eight potential trigger points, six of which have justification. After these points, however, the agent continues in its original direction. This case provides partial evidence for a possible local mechanism through which timely prompts alter subsequent actions near a loop.

Repetition in a discrete action structure and sustained similarity among actions should be modeled separately. The former is better suited to sequence-pattern detection with semantic information, whereas the latter requires jointly considering continuous similarity and task progress. Treating both phenomena uniformly as superficial repetition is likely to produce both false negatives and false positives.

Natural-language interventions can help reduce tool-use errors. Among trajectories containing at least one Failed Action, the mean number of Failed Action occurrences for DeepSeek-V4-Pro decreases by 5.7--12.8\% across the three strategies. For Qwen3-Coder-Plus, it decreases by 5.5\% under Sequitur and 8.8\% under Matrix Profile, while Semantic Sequitur shows a small increase of 3.0\%. Some configurations nevertheless show higher within-run repetition of individual tool–error pairs, suggesting that lower mean counts do not necessarily imply effective recovery from recurring errors.

Frequent, generic reminders risk disrupting otherwise reasonable recovery. In a missing-pytest case, the agent in baseline installs the dependency subsequently, whereas the agent in intervention group first attempts to write alternative code and returns to installation only after additional occurrences of failure. A tool failure is therefore not inherently an anomaly that requires reflection. Failures may arise from normal error reproduction, hypothesis testing, or environmental problems that can be repaired directly. Failed Action occurs frequently, repeated prompts asking the agent to reconsider error causes may interrupt appropriate corrective actions and introduce new tool-use detours. Triggering conditions should therefore consider error recurrence and the agent’s existing recovery efforts.

Repeated short sequences do not lead to custom tool reuse. Fewer than five Custom Tool Candidate detections occur in total, and no custom tools are created or invoked. The identified candidate sequences typically resemble \texttt{bash-python} followed by \texttt{str\_replace\_editor} and then \texttt{bash-python}. They have little functional boundaries and reuse value. The current detection approach based on repetitions of short sequences cannot reliably identify tool workflows suitable for encapsulation from bug-fix trajectories. One possible reason is that bug-fix processes are highly context dependent. Code browsing is concentrated primarily in the early stages, after which agents dynamically alternate between modification and validation. Similar tool sequences therefore do not necessarily serve the same function. \textbf{Finding 3. Mechanical pattern matching alone is insufficient, and semantic judgments can help or mislead. Despite strong context dependence, the pilot results support the effectiveness of natural-language interventions in improving tool use of agents.}

Overall, the pilot study provides preliminary evidence that online natural-language interventions can improve task-oriented agent's tool-use behavior and reduce interaction costs under some conditions. These findings motivate our work’s focus on tracking execution context and providing selective, timely assistance.

\section{Online Assistance for Agentic Tooling: The Method}\label{sec4}
\subsection{Why Cueing and Spotting}

As the pilot study illustrates, agents use tools through context-dependent exploration, feedback interpretation, and recovery, calling for selective assistance that supports their autonomous adjustment as tasks unfold. We therefore use linguistic cues and context-specific guidance to help agents interpret execution evidence and revise their tool use, avoiding mechanical interventions that treat every repeated action or execution error as a problem. In problem solving, the agent remains responsible for understanding the task, selecting tools, interpreting feedback, and planning subsequent actions. Consider a lifter working with a spotter: the lifter controls the movement and bears the weight, while a spotter observes and assists when needed. Likewise, runtime support should allow the agent to respond to tool feedback independently, intervening when a local action is ill-suited to the current situation or a sequence of interactions develops into a detour. In exercise instruction, cueing and spotting are commonly delivered through verbal, visual, and tactile modalities.\footnote{The National Academy of Sports Medicine (NASM) includes "kinesthetic, auditory, and visual cueing techniques" in its Certified Personal Trainer (CPT) exam blueprint under exercise technique and training instruction, which can be discussed as verbal, visual, and tactile supports. See \url{https://www.nasm.org/docs/nasmlibraries/pdf/nasm-cpt-exam-blueprint.pdf}.} A spotter may verbally prompt the lifter to adjust a movement, use a demonstration or visual signal to make a problem easier to notice, or provide physical assistance at a difficult point in the lift. Drawing on this analogy, we provide agents with three forms of assistance: verbal cueing, visual cueing, and tactile spotting.

Verbal cueing provides lightweight support through brief natural language messages that draw attention to a problematic interaction or suggest an adjustment. Its effectiveness depends on the agent's interpretation: generic reminders may offer little useful information, while misguided cues may disrupt reasonable actions. Visual cueing addresses information presentation. Tool calls and their outcomes are often scattered across a long trajectory, making consequential anomalies easy to overlook. Reorganizing and emphasizing relevant information helps the agent recognize the relationship among an action, its feedback, and a potential problem. Tactile spotting provides context-specific guidance, analogous to a spotter assisting at a difficult point in a lift. It offers a concrete, feasible direction for adjustment while the agent retains responsibility for carrying out the task.

Delivering these forms of assistance during task execution requires determining when support is warranted and what it should address. Tailoring interventions to local clumsy tool-use behaviors and persistent tool-use detours requires tracking the agent's evolving execution state and interaction patterns. The detection of clumsy tool-use behaviors and tool-use detours must account for differences in temporal scope and execution context rather than rely on a single uniform scale. These requirements motivate the design of a middleware support layer for autonomous tool use that connects symptom detection with selective, context-sensitive cueing and spotting.

\subsection{ToolPulse}

\begin{figure}
	\centering
	\includegraphics[
        width=\linewidth,
        height=0.70\textheight,
        keepaspectratio
    ]{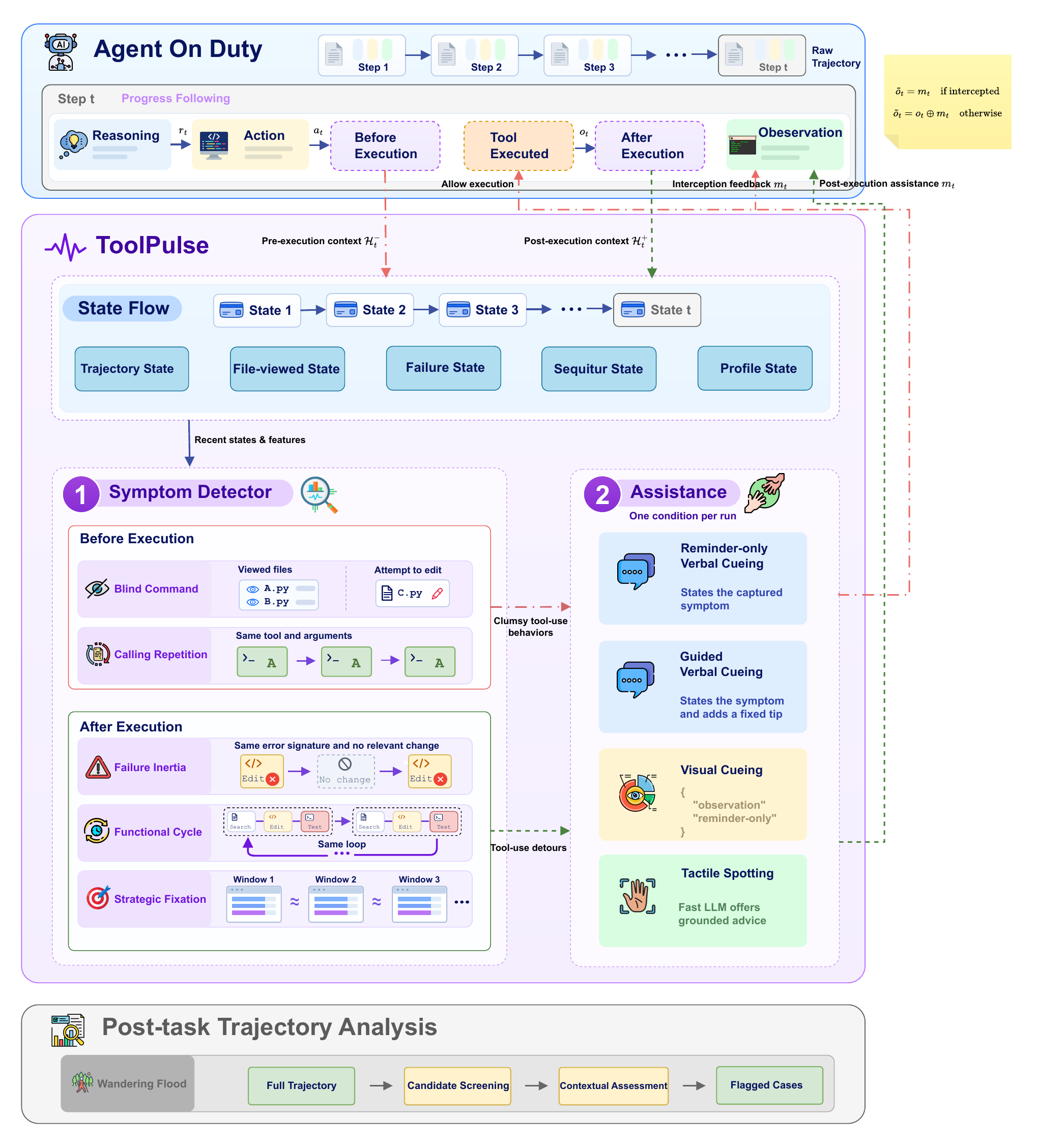}
    \caption{ToolPulse overview. At step $t$, $r_t$, $a_t$, and $o_t$ denote the agent's reasoning record, proposed tool call, and execution result, respectively. The pre-execution context $\mathcal H_t^-$ comprises the task input, prior trajectory, $r_t$, and $a_t$; the post-execution context $\mathcal H_t^+$ adds $o_t$. Red and green dashed arrows mark the respective pathways. ToolPulse tracks the evolving state flow, detects clumsy tool-use behaviors and tool-use detours, and provides assistance $m_t$ under one of four conditions (one per run). The agent receives $\widetilde{o}_t=m_t$ for intercepted calls or $\widetilde{o}_t=o_t\oplus m_t$ otherwise, where $\oplus$ denotes joint presentation of tool output and assistance, and remains responsible for subsequent decisions and actions. Wandering Flood is assessed after task completion through candidate screening and contextual assessment of the complete trajectory, without triggering online assistance.}
    \label{fig:toolpulse}
\end{figure}

Building on the pilot study in Section~\ref{sec3} and the design considerations above, we introduce ToolPulse, a state-flow-tracking online tool-use assistance middleware that supports autonomous tool use by tracking an agent's evolving tool interactions and providing timely support when needed. Fig~\ref{fig:toolpulse} shows the overview. Embedded in the agent's execution process, it closely follows each tool interaction and updates a state flow consisting of the proposed action, execution feedback, and recently observed states. Let $q$ denote the task input and $\tau$ the interaction trajectory defined previously. At step $t$, the agent produces a reasoning record $r_t$ and proposes an action $a_t=(u_t,p_t)$, where $u_t$ is the tool name and $p_t$ contains its arguments. ToolPulse has two observation points: after the action is proposed but before it is executed, and after the tool returns but before the agent resumes reasoning. The information available at these points is
\begin{equation}
\mathcal H_t^-=(q,\tau_{<t},r_t,a_t),
\qquad
\mathcal H_t^+=(\mathcal H_t^-,o_t)
\end{equation}
where $o_t$ is the tool execution result. When ToolPulse intercepts the proposed action before execution, no tool execution result is produced, and we set $o_t=\bot$. ToolPulse uses $\mathcal H_t^-$ to inspect proposed calls and, when necessary, intercept operations that lack sufficient grounding in the observed environment. It uses $\mathcal H_t^+$ to assess the completed local interaction and the recent trajectory. Let $m_t$ contain all assistance generated at step $t$, and let $\widetilde{o}_t$ be the observation delivered to the agent. Then
\begin{equation}
\widetilde{o}_t=
\begin{cases}
m_t, & o_t=\bot\\
o_t\oplus m_t, & o_t\neq\bot
\end{cases}
\end{equation}
where $\oplus$ presents the original tool execution result together with the assistance. When no assistance is triggered, $m_t=\varnothing$ and $\widetilde{o}_t=o_t$. If multiple symptoms are captured at the same step, $m_t$ contains their corresponding messages. By collecting evidence close to the point at which a behavior occurs and providing targeted assistance only when a relevant symptom is detected, ToolPulse limits the overhead associated with continual reflection or global replanning while preserving the agent's execution process.

An agent's tool use unfolds through successive interactions with the environment in pursuit of a task goal. When solving a coding task, an agent typically moves between exploration and progress. It explores to locate relevant code, gather information about the environment, and narrow the problem scope. It then uses that information to modify the program, test hypotheses, and advance the task. New evidence may require it to return to exploration. Throughout this process, the agent must remain grounded in the actual program state and tool feedback. When a tool call fails or a strategy ceases to work, it must also recover by diagnosing the failure and adjusting its subsequent actions.

We now extend the pilot study's analysis of individual actions and recurring interaction patterns to characterize tool-use difficulties at two temporal granularities. The first concerns relatively well-bounded local interactions, including an individual action or a short sequence of adjacent actions. We call behaviors that are inconsistent with the current task state, insufficiently constrained, or clearly redundant clumsy tool-use behaviors. They can generally be assessed from limited context immediately before or after the relevant action. The second granularity concerns longer segments of consecutive actions. Each action may be individually defensible, yet the sequence may collectively fail to advance the task. We call a persistent interaction pattern that leads the agent away from productive problem solving a tool-use detour. Detecting a detour requires examining relationships among recent actions, changes in state, and whether the agent has made a substantive strategic adjustment.

Studying these difficulties empirically requires identifying concrete symptoms of local missteps and persistent detours. We therefore draw on the pilot study in Section~\ref{sec3} to organize three clumsy tool-use behaviors and three tool-use detours along four dimensions: exploration, progress, grounding, and recovery. Table~\ref{tab:toolpulse_symptoms} summarizes the symptoms. We combine temporal detection, state-based rules, and semantic judgments at multiple granularities. For bug-fix tasks, we implement this approach as follows.

\begin{table}[t]
\caption{Clumsy tool-use behaviors and tool-use detours considered in ToolPulse. Observations of repeated calls, recurring action patterns, and repeated failures in the pilot study (Section~\ref{sec3}) informed this organization and motivated closer attention to exploration scope, task progress, action grounding, and recovery efforts.}
\label{tab:toolpulse_symptoms}
\centering
\small
\begin{tabular*}{\linewidth}{@{\extracolsep{\fill}}
p{0.13\linewidth}
p{0.18\linewidth}
p{0.24\linewidth}
p{0.35\linewidth}
@{}}
\toprule
\textbf{Dimension} & \textbf{Symptom} & \textbf{Category} & \textbf{Description} \\
\midrule
Exploration
& Wandering Flood
& Clumsy tool-use behavior
& Excessive exploration whose scope is insufficiently constrained and may incur substantial hidden costs. \\

\addlinespace
Progress
& Calling Repetition
& Clumsy tool-use behavior
& Consecutive calls with identical tool names and arguments. \\

Progress
& Functional Cycle
& Tool-use detour
& Periodic repetition of an ordered sequence of actions with the same semantic functions. \\

Progress
& Strategic Fixation
& Tool-use detour
& Sustained similarity in multistep interaction features within a short sliding window. \\

\addlinespace
Grounding
& Blind Command
& Clumsy tool-use behavior
& A potentially risky action taken before the relevant environment state has been established through observation. \\

\addlinespace
Recovery
& Failure Inertia
& Tool-use detour
& Repeated tool failures without a substantive recovery action. \\
\bottomrule
\end{tabular*}
\end{table}

Calling Repetition, Blind Command, and Failure Inertia are captured by state-based rules together with semantic information. Calling Repetition is detected when three consecutive proposed actions have identical tool names and arguments:
\begin{equation}
\mathrm{CR}_t=
\mathbf{1}\!\left[t\geq 3
\land a_{t-2}=a_{t-1}=a_t\right]
\end{equation}

For Blind Command, let $V_{t-1}$ contain the files whose contents have been observed through a view-related action before step $t$. Writing $\operatorname{View}(a_i,f)$ for an action and result that reveal the contents of file $f$, we define
\begin{equation}
V_{t-1}
=
\{f:\exists i<t,\operatorname{View}(a_i,f)\},
\qquad
\mathrm{BC}_t
=
\mathbf{1}\!\left[
\operatorname{Edit}(a_t,f)
\land f\notin V_{t-1}
\right]
\end{equation}

For a failed action at step $t$, define its error signature as
\begin{equation}
\sigma_t
=
\bigl(u_t,\operatorname{errorType}_t,
\operatorname{errorFingerprint}_t\bigr)
\end{equation}
Let $i$ be the most recent earlier failed call within a short lookback window, and let $z_j$ be the semantic label of action $a_j$. We use $\mathcal L_{\mathrm{change}}$ to denote labels associated with changes to the program or environment. Failure Inertia is detected when the two failed calls have the same error signature and no such change occurs between them:
\begin{equation}
\mathrm{FI}_t
=
\mathbf{1}\!\left[
\sigma_i=\sigma_t
\land
\nexists j\in(i,t):
z_j\in\mathcal L_{\mathrm{change}}
\right]
\end{equation}

For Functional Cycle, each tool call is mapped to a bug-fix-oriented semantic label in Table~\ref{tab:bugfix_labels}, and the label sequence is processed by Sequitur. Let $G_t=\operatorname{Sequitur}(z_{1:t})$, and let $\operatorname{Rules}(G_t)$ denote the label sequences obtained by expanding its grammar rules. A cycle is detected when a sequence of more than one label occurs three consecutive times at the end of the current trajectory:
\begin{equation}
\mathrm{FC}_t
=
\mathbf{1}\!\left[
\exists s\in\operatorname{Rules}(G_t):
|s|>2,\;
t\geq 3|s|,\;
z_{t-3|s|+1:t}=s^3
\right]
\end{equation}
where $s^3$ denotes three consecutive copies of $s$. To avoid repeated interventions for the same cycle, each distinct semantic label sequence triggers assistance at most once within a task trajectory.

For Strategic Fixation, each interaction is encoded as a feature vector $x_t$ containing the length of the agent's reasoning record, hashes of the tool name and arguments, the observation length, execution time, and an error indicator. The latest $3w$ steps are divided into three consecutive, non-overlapping windows:
\begin{equation}
W_1=x_{t-3w+1:t-2w},\qquad
W_2=x_{t-2w+1:t-w},\qquad
W_3=x_{t-w+1:t}
\end{equation}
Let $d_{\mathrm{MP}}$ denote the window distance obtained using Matrix Profile. Strategic Fixation is detected when three pairs of adjacent windows fall below the threshold $\theta_S$:
\begin{equation}
\mathrm{SF}_t
=
\mathbf{1}\!\left[
d_{\mathrm{MP}}(W_1,W_2)<\theta_S
\land
d_{\mathrm{MP}}(W_2,W_3)<\theta_S
\land
d_{\mathrm{MP}}(W_1,W_3)<\theta_S
\right]
\end{equation}

Wandering Flood is more dependent on task context and is difficult to capture reliably with a deterministic online rule. A broad search may produce substantial output yet still be justified when the task requires understanding global call relationships. We therefore do not detect or cue Wandering Flood during execution. Instead, we analyze it from the complete trajectory after the task ends, avoiding interventions based on incomplete rules that could disrupt legitimate exploration.

We propose a set of runtime assistance strategies that connect symptom detection with cueing and spotting to help agents recognize tool-use difficulties and adjust their subsequent actions. Reminder-only verbal cueing states the observed behavior without prescribing an adjustment. Guided verbal cueing adds a brief, fixed direction for improvement. Visual cueing changes the organization of the feedback. Motivated by prior evidence that structured representations may affect LLM reasoning efficiency and information transfer \citep{chen-etal-2024-beyond-natural}, it places the original observation and the reminder-only verbal message in a JSON object, changing only how the environment feedback and behavioral reminder are structured relative to one another. Tactile spotting uses a fast language model to generate advice grounded in the current interaction and directed toward a feasible next action. Verbal cueing states the problem, visual cueing changes its presentation, and tactile spotting provides context-specific guidance through additional inference. In all four strategies, the agent makes the subsequent decision and carries out the task. We study these strategies in a relatively controlled setting that reflects engineering practice. We will further present a reference implementation informed by these observations, providing a starting point for integrating ToolPulse into agent execution environments.

\section{Evaluation and Analysis}\label{sec5}
\subsection{Study Design}
We evaluate ToolPulse as runtime support for autonomous tool use, focusing on whether its assistance helps agents recognize tool-use difficulties, make task-relevant adjustments, and sustain task progress. The evaluation connects interaction-level behavior with task-level outcomes. At the interaction level, we examine whether detected symptoms are supported by the execution context and whether agents respond with substantive corrections. At the task level, we assess patch submission, successful repair, and resource expenditure. This complements the goal-oriented evaluation commonly used in agent benchmarks \citep{jimenez2024swebench,liu2025agentbenchevaluatingllmsagents} with evidence about how agents adjust their tool use during execution.

Evaluating such support requires interpreting behavior in its execution context and across time. Repeated calls can serve a legitimate investigative purpose, and an execution error can be part of intentional failure reproduction. Local interactions and persistent detours also require different scopes of evidence. We therefore assess symptoms using criteria suited to their temporal scope and distinguish detection validity, behavioral cessation, and substantive corrective responses. This distinction connects an observed change in action to whether the agent addresses the underlying difficulty. Task outcomes provide a further perspective on the practical consequences of these adjustments.

We compare an unassisted baseline with four assistance strategies: reminder-only verbal cueing, guided verbal cueing, visual cueing, and tactile spotting. The baseline follows the original SWE-agent workflow without ToolPulse monitoring or additional assistance. The four assistance strategies share the same detection settings for Calling Repetition, Functional Cycle, Strategic Fixation, Blind Command, and Failure Inertia, and differ in the assistance delivered after a trigger. Wandering Flood is assessed offline after task completion, providing complementary evidence about exploration quality across the evaluated trajectories.

The assistance conditions support three targeted comparisons. Guided versus reminder-only verbal cueing examines the contribution of an explicit adjustment direction. Visual versus reminder-only verbal cueing examines feedback organization while retaining the same reminder content. Tactile spotting versus guided verbal cueing examines the contribution of context-specific, model-generated advice. Together, these comparisons examine how the content and presentation of runtime support relate to agents' subsequent behavioral adjustments.

We retain the two pilot backbones for continuity and add a compact, open-weight model to examine variation across model settings. With the agent framework and tool environment held fixed, the experiment crosses three backbone models with three programming languages. We evaluate 90 bug-fix tasks, comprising 30 tasks each in Python, C++, and Java. Each model--task--condition combination is run once, yielding $3\times90\times5=1{,}350$ trajectories: 270 baseline trajectories and 1,080 assisted trajectories. Within each backbone model, all conditions use the same task instances, enabling task-level comparisons to be paired by instance. The experiment reuses the 60 Python baseline trajectories for DeepSeek-V4-Pro and Qwen3-Coder-Plus from the pilot study. These runs use the same model versions, tool environment, and final evaluation procedure as the main experiment. The remaining 1,290 trajectories are generated for the main experiment.

We then analyze the following research questions in detail:

\textbf{RQ1)} \textit{To what extent can ToolPulse tackle clumsy tool-use behaviors and tool-use detours?}

\textbf{RQ2)} \textit{How do different forms of cueing and spotting affect clumsy tool-use behaviors and tool-use detours?}

\textbf{RQ3)} \textit{To what extent can ToolPulse support task completion, and what costs and limitations accompany its use?}

\subsection{Experimental Configuration}

\paragraph{Agent and Models.}
We continue to use SWE-agent as the base agent framework. We retain DeepSeek-V4-Pro and Qwen3-Coder-Plus as backbone models and introduce Qwen3.6-27B \citep{qwen3.6-27b}, a 27B dense model released by the Qwen team with BF16 weights. Compared with the other two models, it provides a more compact, open-weight model, allowing us to further examine whether the effect of runtime assistance depends on the size of backbone model. All experiments use the same SWE-agent interaction interface and tool environment. Tactile spotting uses Qwen3.7-Flash \citep{qwen37} as a fast auxiliary model to generate context-specific advice from the current interaction context at runtime.

\paragraph{Task.}
We continue to conduct experiments on the bug-fix subset of OmniCode. Whereas the pilot study included only Python tasks, we extend the scope to Python, Java, and C++. We also retain the complexity-stratified sampling method used in the pilot study. Complexity rankings are computed independently within each language, and 30 instances are sampled from each language, yielding 90 bug-fix tasks in total.

\paragraph{Evaluation Measures.}
We evaluate the results at two levels: task outcomes and behavioral quality.

For task outcomes, we count the tasks for which the agent successfully submits a patch, the tasks for which the patch passes the tests, and token usage. Token usage is the sum of the backbone model's input and output tokens over the complete trajectory. For assistance strategy $a$ and its set of instances $\mathcal I$ paired with the baseline, we compute $\Delta_T(\mathcal I,a)$ using Eq.~\eqref{eq:pilot_token_change}, with $a$ replacing $s$. Here, $T_{a,i}$ includes the input and output tokens of both the backbone model and any auxiliary model used at runtime on instance $i$.

We examine behavioral quality as a process-oriented dimension of autonomous tool use. The stochastic nature of agent execution and the dependence of individual actions on evolving task context make this quality difficult to capture through a single aggregate metric. We therefore operationalize its assessment through observable tool-use symptoms and judgments grounded in the surrounding trajectory. For Calling Repetition, Functional Cycle, Strategic Fixation, Blind Command, and Failure Inertia, we record symptom occurrences and assess trigger validity, subsequent behavioral cessation, and substantive corrective responses.

Assessing trigger validity and subsequent adjustment for Functional Cycle, Strategic Fixation, and Failure Inertia requires examining multiple interaction steps. We therefore use Kimi-K3 \citep{kimiteam2026kimik3openfrontier} as an LLM judge. For each trigger, the judge receives the task description, the prompt shown to the agent, up to eight steps before the trigger, and up to three steps after it. The longer pre-trigger window provides context for assessing how the detected pattern developed, while the shorter post-trigger window focuses the assessment on the agent's immediate response to assistance. It outputs \texttt{trigger\_validity} (\texttt{valid}, \texttt{invalid}, or \texttt{unclear}), \texttt{behavior\_after\_trigger} (\texttt{persists}, \texttt{ceases}, or \texttt{unclear}), \texttt{corrective\_response} (\texttt{substantive}, \texttt{non\_substantive}, or \texttt{unclear}), \texttt{evidence\_step\_ids}, and \texttt{brief\_reason}. The last two fields record the steps supporting the judgment and a brief rationale.

Calling Repetition and Blind Command have more explicit operational criteria. For Calling Repetition, consecutive repeated triggers within the same instance are merged into one episode. If the agent immediately switches to a different action after the trigger, \texttt{behavior\_after\_trigger} is assigned \texttt{ceases}. If repetition continues for several steps before stopping, it is assigned \texttt{persists}. If repetition continues until the trajectory ends, it is assigned \texttt{unclear}. If a new, nonconsecutive repetition episode occurs later in the same instance, \texttt{corrective\_response} is assigned \texttt{non\_substantive}; otherwise, it is assigned \texttt{substantive}. For Blind Command, checking the target file before editing it again, or checking it without subsequently editing it, maps to \texttt{ceases} and \texttt{substantive}. Editing the file again without first checking it maps to \texttt{persists} and \texttt{non\_substantive}. Neither checking nor editing it afterward maps to \texttt{ceases} and \texttt{non\_substantive}. Thus, the rule-assigned trigger validity for these two categories is distinguished from the judge-assessed validity of the other three.

Wandering Flood can expand exploration without a corresponding gain in task-relevant evidence, making it harder for the agent to narrow the problem scope and choose a productive next action. It is identified through a two-stage process of low-cost screening followed by LLM judgment, which is inspired by prior work \citep{sinha2024realtimeanomalydetectionreactive,li-etal-2026-towards}. In the first stage, we examine the Retrieve actions listed in Table~\ref{tab:bugfix_labels}. For each language, we separately aggregate the observation lengths and tool execution times of Retrieve-type actions across the same 30 tasks for each backbone model, and calculate their 95th percentiles as screening thresholds. We then apply the thresholds for each backbone model to the corresponding task set. A step becomes a candidate if either its observation length or its tool execution time exceeds the respective threshold. In the second stage, we provide the candidate step, up to eight preceding steps, and up to three subsequent steps to an LLM judge. The judge determines whether the step constitutes Wandering Flood and outputs \texttt{verdict} (\texttt{yes}, \texttt{no}, or \texttt{uncertain}), \texttt{information\_gain} (\texttt{sufficient}, \texttt{limited}, or \texttt{unclear}), and \texttt{rationale}, which records a brief reason for the judgment. To check for potential missed cases, we also sample at most two steps per instance that did not pass the first-stage screen but whose measures fall between the 90th and 95th percentiles, and assess them using the same judge. 

The complete prompt templates, task list, run configurations, and details of the manual verification of the LLM judge's outputs are provided in Appendix~\ref{secA2}.

\subsection{RQ1}\label{secr1}

 We analyze the 1,080 trajectories produced under the four assistance strategies to assess ToolPulse's ability to tackle clumsy tool-use behaviors and tool-use detours. For the five symptoms captured online, we examine detection validity, post-assistance cessation, and substantive corrective responses. We additionally assess Wandering Flood offline across all 1,350 trajectories and examine symptom occurrence rates across backbone models, programming languages, and task difficulty levels.

\textbf{ToolPulse has the ability to identify contextually supported clumsy tool-use behaviors and tool-use detours across the evaluated trajectories.} 
As shown in Table~\ref{tab:symptom_outcomes}, Strategic Fixation is detected in 570 of the 1,080 assisted trajectories, with 1,134 of its 1,841 triggers judged valid. Functional Cycle and Failure Inertia account for 29 of 48 and 110 of 154 valid triggers, respectively. Together with the rule-defined Calling Repetition and Blind Command events, these results show that ToolPulse captures both local tool-use problems and longer interaction patterns that warrant corrective attention.

\begin{table}[t]
\caption{Online clumsy tool-use behaviors and tool-use detours in 1,080 assisted trajectories. \textit{Affected Trajectory} counts runs with at least one trigger; \textit{Event} counts step-level triggers. \textit{Valid} is judge-valid triggers over all triggers (rule-defined for Calling Repetition and Blind Command). \textit{Ceased} is stopped over valid triggers, or over all triggers for the two rule-defined categories. \textit{Substantive Response} is substantive over ceased events. Discrepancies between totals and category counts reflect cases judged \textit{unclear}}
\label{tab:symptom_outcomes}
\centering
\small
\resizebox{\linewidth}{!}{%
\begin{tabular}{lccccc}
\toprule
\textbf{Symptom}
& \textbf{Affected Trajectory}
& \textbf{Event}
& \textbf{Valid}
& \textbf{Ceased}
& \textbf{Substantive Response} \\
\midrule

Calling Repetition
& 34 / 1,080
& 164
& Rule-defined
& 80 / 164 (84 persists)
& 20 / 80 (60 non-substantive) \\

Functional Cycle
& 42 / 1,080
& 48
& 29 / 48 (19 invalid)
& 19 / 29 (10 persists)
& 19 / 19 (0 non-substantive) \\

Strategic Fixation
& 570 / 1,080
& 1,841
& 1,134 / 1,841 (700 invalid)
& 715 / 1,134 (417 persists)
& 699 / 715 (16 non-substantive) \\

Blind Command
& 36 / 1,080
& 86
& Rule-defined
& 67 / 86 (19 persists)
& 41 / 67 (26 non-substantive) \\

Failure Inertia
& 101 / 1,080
& 154
& 110 / 154 (42 invalid)
& 100 / 110 (10 persists)
& 100 / 100 (0 non-substantive) \\

\bottomrule
\end{tabular}%
}
\end{table}

\textbf{Agents often make substantive adjustments after receiving ToolPulse assistance for tool-use detours.} 
Substantive responses occur after 19 of 29 valid Functional Cycle triggers (65.5\%), 699 of 1,134 valid Strategic Fixation triggers (61.6\%), and 100 of 110 valid Failure Inertia triggers (90.9\%). Moreover, cessation generally corresponds to a substantive adjustment: all ceased Functional Cycle and Failure Inertia events receive such a response, as do 699 of the 715 ceased Strategic Fixation events (97.8\%). These responses indicate that agents often move beyond merely interrupting a detected pattern and take corrective action within the observed post-trigger window. Failure Inertia shows the highest response proportion, indicating that repeated failures accompanied by an absence of recovery actions provide particularly actionable intervention opportunities.

\textbf{Local clumsy tool-use behaviors also respond to assistance, although the durability of correction varies.} 
Blind Command ceases after 67 of 86 triggers (77.9\%), and 41 of these cessation events involve a substantive response: the agent inspects the target file before a subsequent edit or inspects it without editing it again. Calling Repetition ceases after 80 of 164 triggers (48.8\%). Of these 80 events, 20 satisfy the substantive-response criterion, whereas the remaining 60 are followed by another repetition episode later in the task. ToolPulse therefore supports immediate correction of both local symptoms, while sustained avoidance of repeated calls remains more difficult.

\begin{figure}
	\centering
	\includegraphics[width=.9\textwidth]{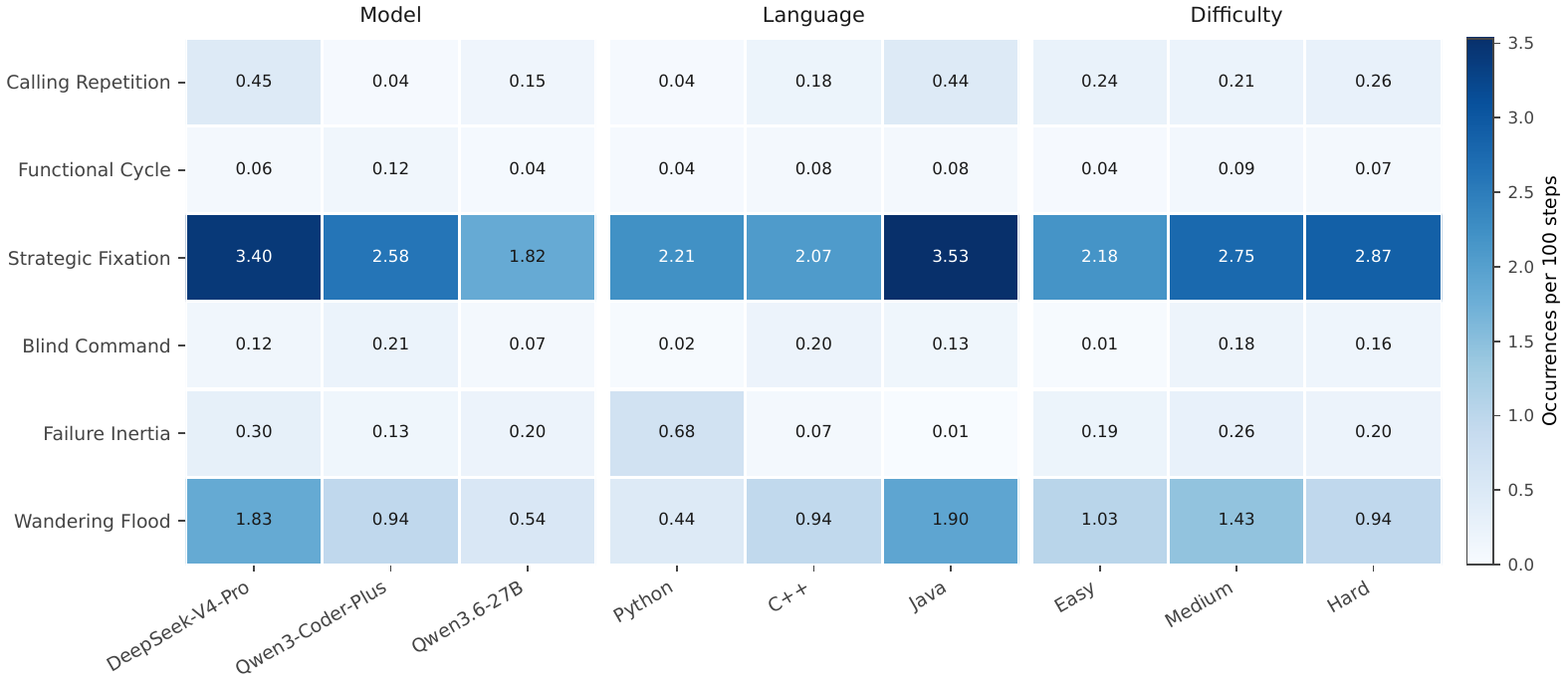}
    \caption{Occurrence rates of six tool-use symptoms per 100 agent steps, grouped by backbone model, programming language, and task difficulty. Darker cells indicate higher occurrence rates.}
    \label{fig:category_heatmap}
\end{figure}

\textbf{Symptom profiles vary across task difficulty levels, languages and models.}
Figure~\ref{fig:category_heatmap} reports occurrences per 100 agent steps. Strategic Fixation is the most frequent symptom in every reported group. Its rate is 3.40 with DeepSeek-V4-Pro and 1.82 with Qwen3.6-27B, and reaches 3.53 in Java, compared with 2.21 in Python and 2.07 in C++. Its rate also increases from 2.18 on easy tasks to 2.75 on medium tasks and 2.87 on hard tasks. Failure Inertia follows a different distribution, occurring 0.68 times per 100 steps in Python, compared with 0.07 in C++ and 0.01 in Java. These differences show that the relative prominence of clumsy tool-use behaviors and tool-use detours depends on the execution setting.

\textbf{Excessive exploration is a recurring tool-use difficulty across execution settings.}
The first-stage screen selects 3,191 candidate steps from 935 trajectories. Subsequent assessment identifies 799 Wandering Flood events across 400 of the 1,350 trajectories (29.6\%; Table~\ref{tab:wandering_flood}). Its occurrence rate is higher with DeepSeek-V4-Pro than with Qwen3.6-27B (1.83 versus 0.54 per 100 steps), and higher in Java than in Python (1.90 versus 0.44). The rate peaks on medium-difficulty tasks (1.43), exceeding that on hard tasks (0.94). These results establish excessive exploration as a recurring issue whose frequency does not simply increase with task difficulty. The distinction between confirmed events and screened candidates also demonstrates the importance of assessing exploration in context, because broad retrieval can remain justified by the task.

\begin{table}[t]
\caption{Offline Wandering Flood analysis across 1,350 trajectories. \textit{Candidate Trajectory} counts runs with at least one screened step; \textit{Event} counts candidates judged. \textit{Verdict} is positive over judged candidates. \textit{Positive Trajectory} counts runs with at least one positive step. Discrepancies between totals and category counts reflect cases judged \textit{unclear}}
\label{tab:wandering_flood}
\centering
\small
\begin{tabular*}{\linewidth}{@{\extracolsep{\fill}}lcccc@{}}
\toprule
\textbf{Symptom}
& \textbf{Candidate Trajectory}
& \textbf{Event}
& \textbf{Verdict}
& \textbf{Positive Trajectory} \\
\midrule

Wandering Flood
& 935 / 1,350
& 3,191 candidates judged
& 799 / 3,191 (2,391 no)
& 400 / 1,350 \\

\bottomrule
\end{tabular*}
\end{table}

\begin{rqbox}
\textbf{Answer to RQ1:}

The results support ToolPulse's ability to identify clumsy tool-use behaviors and tool-use detours. Substantive corrective responses following assistance further support its role in helping agents adjust their tool use. The extent of adjustment varies by symptom: detour cessation generally involves substantive adjustment, whereas Calling Repetition is more prone to recurrence. These findings support targeted online assistance as a means of helping agents adjust their tool use and recover from unproductive interaction patterns during autonomous task execution. Wandering Flood provides complementary offline evidence of excessive exploration.
\end{rqbox}

\subsection{RQ2}\label{secr2}

Building on RQ1, we compare all four assistance strategies through substantive-response rates and subsequent trigger patterns. We first examine their aggregate response profiles across the five online symptoms, followed by three targeted comparisons. Guided versus reminder-only verbal cueing examines whether specifying an adjustment direction adds value beyond identifying the problem. Visual versus reminder-only verbal cueing examines feedback organization while retaining the same reminder content. Tactile spotting versus guided verbal cueing examines whether context-specific, model-generated advice provides further benefit beyond fixed guidance.

\textbf{Concise reminders provide the strongest aggregate support for substantive adjustment.}
Pooling substantive responses and eligible triggers across the five symptoms, reminder-only verbal cueing achieves a response rate of 67.8\% (158/233), compared with 59.2\% (141/238) for guided verbal cueing, 52.6\% (152/289) for visual cueing, and 56.1\% (428/763) for tactile spotting (Table~\ref{tab:assistance_outcomes}). It also has the highest response rates for Calling Repetition (25.0\%) and Strategic Fixation (69.8\%). Across the observed mix of eligible events, a brief behavioral reminder therefore provides the most effective form of support within ToolPulse. More elaborate assistance offers advantages for particular symptoms but can also be less effective than concise reminders, as the targeted comparisons below show.

\begin{table}[t]
\caption{For each symptom and assistance condition, each cell reports the trigger count, triggers per 100 attempted agent tool actions in brackets, and substantive corrective responses / eligible triggers. Eligible triggers are rule-defined for Calling Repetition and Blind Command and judged valid otherwise.}
\label{tab:assistance_outcomes}
\centering
\small
\resizebox{\linewidth}{!}{%
\begin{tabular}{lcccc}
\toprule
\textbf{Symptom}
& \textbf{Reminder-Only Verbal Cueing}
& \textbf{Guided Verbal Cueing}
& \textbf{Visual Cueing}
& \textbf{Tactile Spotting} \\
\midrule

Calling Repetition
& 16, [0.09]; 4/16
& 27, [0.15]; 6/27
& 68, [0.38]; 4/68
& 53, [0.30]; 6/53 \\

Functional Cycle
& 12, [0.07]; 7/9
& 15, [0.09]; 6/11
& 15, [0.08]; 6/7
& 6, [0.03]; 0/2 \\

Strategic Fixation
& 236, [1.35]; 118/169
& 236, [1.35]; 96/162
& 245, [1.39]; 106/171
& 1,124, [6.40]; 379/632 \\

Blind Command
& 16, [0.09]; 9/16
& 14, [0.08]; 11/14
& 16, [0.09]; 10/16
& 40, [0.23]; 11/40 \\

Failure Inertia
& 32, [0.18]; 20/23
& 43, [0.25]; 22/24
& 36, [0.20]; 26/27
& 43, [0.24]; 32/36 \\

\bottomrule
\end{tabular}%
}
\end{table}

\textbf{Adding fixed guidance is beneficial when the detected problem has a concrete corrective requirement.}
For Blind Command, guided verbal cueing produces a substantive-response rate of 78.6\% (11/14), compared with 56.3\% (9/16) under reminder-only verbal cueing. Here, the required adjustment is explicit: inspect the target file before editing it again. The higher observed rate is consistent with fixed guidance helping agents connect problem recognition to a corrective action. This advantage does not extend to all symptoms. For Strategic Fixation, both conditions produce 1.35 triggers per 100 attempted agent tool actions, but guided cueing has a substantive-response rate of 59.3\% (96/162), compared with 69.8\% (118/169) under reminder-only cueing. Calling Repetition rates are also similar, at 22.2\% and 25.0\%, respectively. The contribution of fixed guidance thus depends on whether a general instruction adequately specifies the adjustment required by the current interaction.

\textbf{Reorganizing the same reminder changes response patterns without providing a uniform advantage.}
Visual cueing places the original tool observation and the reminder-only message in a JSON object, changing their organization without adding an adjustment direction. Its Failure Inertia substantive-response rate is 96.3\% (26/27), compared with 87.0\% (20/23) under reminder-only cueing. Functional Cycle also has a higher observed rate, 85.7\% (6/7) versus 77.8\% (7/9), although both samples are small. Conversely, Calling Repetition has a lower substantive-response rate, 5.9\% versus 25.0\%, as does Strategic Fixation, 62.0\% versus 69.8\%. Together with the verbal-cueing comparison, these results show that adding a fixed direction and restructuring existing feedback have symptom-dependent contributions. Neither consistently improves on the reminder-only condition.

\textbf{Context-specific advice does not consistently improve on fixed guidance.}
Tactile spotting generates a corrective recommendation from the current interaction context through an auxiliary model. Its Strategic Fixation substantive-response rate is 60.0\%, close to the 59.3\% under guided verbal cueing, while its Failure Inertia rate is 88.9\%, compared with 91.7\%. For Blind Command and Calling Repetition, tactile spotting reaches 27.5\% and 11.3\%, respectively, compared with 78.6\% and 22.2\% under guided cueing. The current spotting configuration therefore does not translate its additional contextual inference into consistently higher substantive-response rates.

The larger number of Strategic Fixation events under tactile spotting primarily reflects consecutive re-triggering.
To interpret its event-level outcomes, we examine how triggers are distributed within trajectories. A consecutive trigger run is a maximal sequence of at least two adjacent agent-action steps that all trigger Strategic Fixation. Guided verbal cueing and tactile spotting affect similar numbers of trajectories, 140 and 144, respectively, but their trigger frequencies are 1.35 and 6.40 per 100 attempted agent tool actions. Under tactile spotting, 107 of the 144 affected trajectories contain consecutive trigger runs, and 974 of the 1,124 triggers (86.7\%) belong to these runs. Run length has a median of 3 and a maximum of 30. None of the three cueing conditions contains such runs (Table~\ref{tab:fixation_runs}). Thus, tactile spotting's larger absolute number of substantive responses occurs within a much larger set of repeatedly triggered events, while its response rate remains close to that of guided verbal cueing.

\begin{table}[t]
\caption{A run is a maximal sequence of at least two consecutive Strategic Fixation triggers. \textit{Trajectory with Run} reports trajectories containing runs / trajectories with any Strategic Fixation trigger. \textit{Trigger in Runs} reports triggers in runs / all Strategic Fixation triggers. Run Length reports the median [Q1, Q3] and maximum.}
\label{tab:fixation_runs}
\centering
\small
\begin{tabular*}{\linewidth}{@{\extracolsep{\fill}}lccc@{}}
\toprule
\textbf{Condition}
& \textbf{Trajectory with Run}
& \textbf{Trigger in Runs}
& \textbf{Run Length} \\
\midrule

Reminder-Only
& 0 / 140
& 0 / 236
& N/A \\

Guided Verbal Cueing
& 0 / 140
& 0 / 236
& N/A \\

Visual Cueing
& 0 / 146
& 0 / 245
& N/A \\

Tactile Spotting
& 107 / 144
& 974 / 1,124
& 3 [2, 5], 30 \\

\bottomrule
\end{tabular*}
\end{table}

\begin{rqbox}
\textbf{Answer to RQ2:}

Concise behavioral reminders can be sufficient to elicit substantive adjustment, while additional guidance or structured feedback offers benefits for particular corrective requirements. The current auxiliary model configuration provides no consistent advantage over simpler cueing despite requiring additional inference. These findings suggest that agents' own capacity for judgment and adjustment is more consequential than the strength of external intervention. Runtime assistance should therefore provide selective cues that help agents recognize difficulties and apply their own reasoning to revise subsequent actions.
\end{rqbox}

\subsection{RQ3}

Following the interaction-level analyses in RQ1 and RQ2, we assess task outcomes and token usage across four assistance conditions and an unassisted baseline on 90 bug-fix instances in Python, C++, and Java, using three backbone models. All runs submit a patch; successful repair is measured by whether it passes the issue-specific tests. Token usage includes input and output tokens from the backbone and, where applicable, the auxiliary model. This assessment connects runtime support for autonomous tool use with repair outcomes and resource costs.

\textbf{Backbone-dependent results distinguish repair capability from the capability of tool use.}
Without assistance, DeepSeek-V4-Pro passes 38 of the 90 instances, Qwen3.6-27B passes 35, and Qwen3-Coder-Plus passes 28. These differences exceed the aggregate changes introduced by any assistance condition. Stronger repair performance does not necessarily coincide with fewer tool-use symptoms: Sec~\ref{secr1} reports a higher Strategic Fixation occurrence rate for DeepSeek-V4-Pro than for Qwen3.6-27B, despite its higher baseline pass count. A backbone can therefore produce more correct patches while still exhibiting tool-use difficulties. Correcting an execution pattern may remove an obstacle to progress, while successful repair also depends on diagnosing and fixing the underlying defect.

\begin{table}[t]
\caption{Task outcome comparison under different conditions. \textit{Task} is the number of bug-fix instances evaluated for each language--LLM pair. \textit{Baseline} is the number of instances whose patch passes the issue-specific tests without assistance. Each assistance cell has the form $P\,(G/L)$: $P$ is the number passing under that condition, $G$ is the number that fails under baseline but passes with assistance (newly passed), and $L$ is the number that passes under baseline but fails with assistance (regressed). All comparisons use the same task instances within each language--LLM pair.}
\label{tab:task_outcomes}
\centering
\small
\resizebox{\linewidth}{!}{%
\begin{tabular}{llcccccc}
\toprule
\textbf{Language}
& \textbf{LLM}
& \textbf{Task}
& \textbf{Baseline}
& \multicolumn{4}{c}{\textbf{Assistance Strategy}} \\
\cmidrule(lr){5-8}
&
&
&
&
\textbf{Reminder-Only Verbal Cueing}
& \textbf{Guided Verbal Cueing}
& \textbf{Visual Cueing}
& \textbf{Tactile Spotting} \\
\midrule

Python
& DeepSeek-V4-Pro
& 30
& 21
& 22 (2/1)
& 22 (2/1)
& 22 (1/0)
& 23 (2/0) \\

& Qwen3-Coder-Plus
& 30
& 19
& 18 (1/2)
& 18 (2/3)
& 17 (0/2)
& 17 (0/2) \\

& Qwen3.6-27B
& 30
& 21
& 20 (1/2)
& 21 (1/1)
& 21 (2/2)
& 22 (2/1) \\

\addlinespace

C++
& DeepSeek-V4-Pro
& 30
& 12
& 12 (1/1)
& 12 (0/0)
& 11 (0/1)
& 12 (1/1) \\

& Qwen3-Coder-Plus
& 30
& 5
& 6 (2/1)
& 7 (2/0)
& 5 (1/1)
& 6 (3/2) \\

& Qwen3.6-27B
& 30
& 10
& 12 (2/0)
& 10 (0/0)
& 10 (0/0)
& 11 (1/0) \\

\addlinespace

Java
& DeepSeek-V4-Pro
& 30
& 5
& 5 (0/0)
& 6 (1/0)
& 5 (1/1)
& 5 (1/1) \\

& Qwen3-Coder-Plus
& 30
& 4
& 5 (2/1)
& 4 (1/1)
& 4 (1/1)
& 4 (1/1) \\

& Qwen3.6-27B
& 30
& 4
& 5 (4/3)
& 4 (2/2)
& 4 (2/2)
& 3 (2/3) \\

\bottomrule
\end{tabular}%
}
\end{table}

\begin{figure}
	\centering
	\includegraphics[width=.9\textwidth]{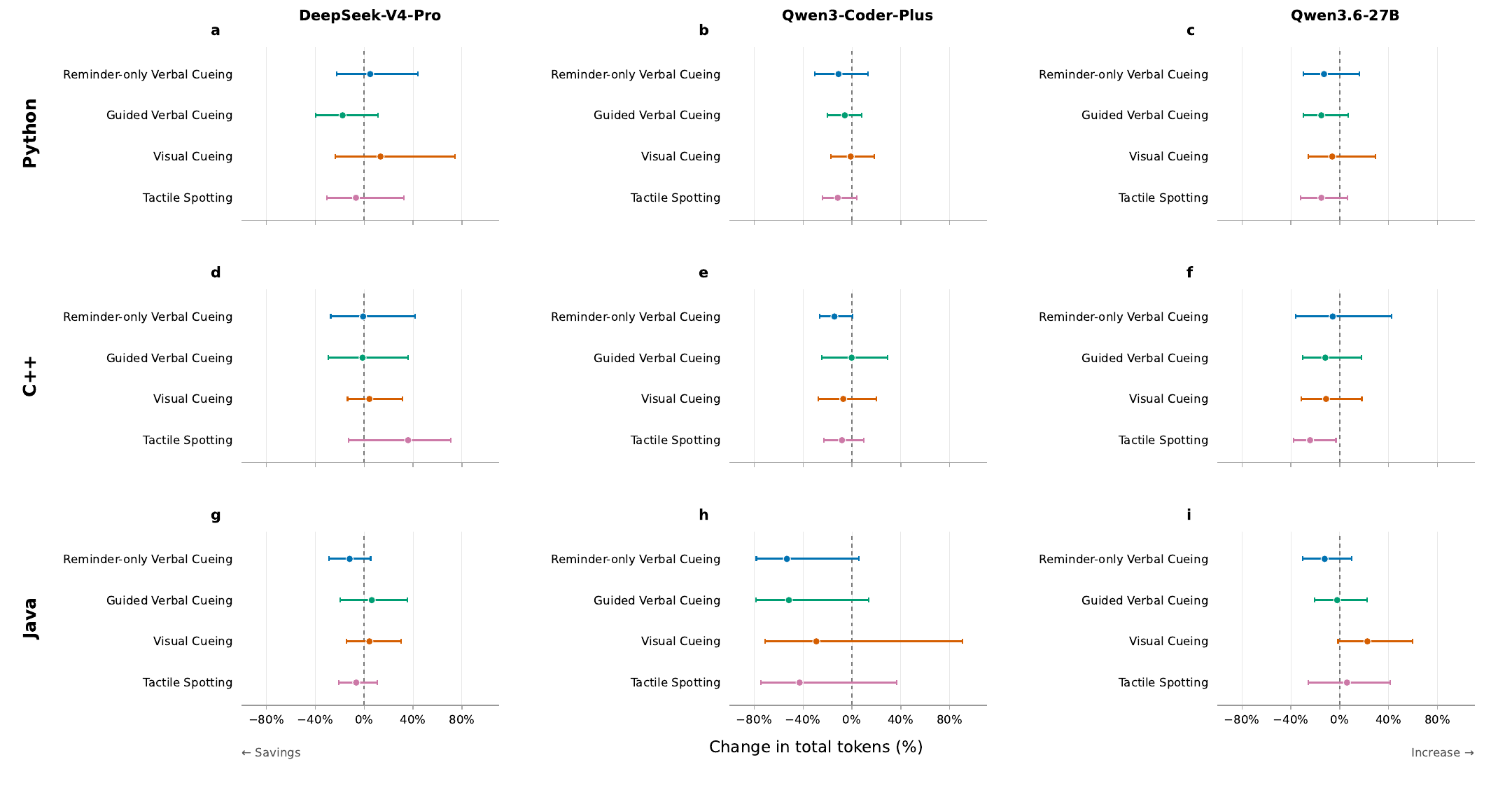}
	\caption{Relative change in total token usage under the four assistance strategies compared with the paired baseline, grouped by programming language and backbone model. Points show percentage changes in summed input and output tokens, including auxiliary-model tokens where applicable. Error bars indicate 95\% paired bootstrap confidence intervals; negative values indicate lower token usage.}
    \label{fig:token_overall}
\end{figure}

\textbf{Runtime support can accompany lower token usage and preserved repair outcomes, with benefits varying across backbone and task settings.}
Figure~\ref{fig:token_overall} reports relative changes in summed token usage over paired instances, with 95\% paired bootstrap confidence intervals. For Qwen3-Coder-Plus on Java, all four assistance conditions have lower token-usage point estimates, with four or five passing patches compared with four under baseline. For tactile spotting with Qwen3.6-27B on C++, the token-change interval lies below zero, while 11 patches pass compared with 10 under baseline. These settings combine lower execution expenditure with retained or increased aggregate repair counts.

Other settings show different trade-offs. For DeepSeek-V4-Pro on C++, tactile spotting has a higher token-usage point estimate while the pass count remains at 12. For reminder-only cueing with Qwen3-Coder-Plus on Python, the token-usage point estimate is lower, but the pass count decreases from 19 to 18. Resource changes must therefore be interpreted alongside repair outcomes and depend on the backbone, task setting, and assistance form. Most confidence intervals include zero, indicating uncertainty in the estimated cost differences.

\textbf{ToolPulse supports behavioral adjustment while broadly preserving task repair outcomes.}
Across the 270 model--task combinations, the baseline produces 101 patches that pass the issue-specific tests, compared with 105 under reminder-only verbal cueing, 104 under guided verbal cueing, 99 under visual cueing, and 103 under tactile spotting (Table~\ref{tab:task_outcomes}). Of the 36 comparisons within language and model settings, 16 leave the pass count unchanged, 13 increase it, and 7 decrease it; no difference exceeds 2 instances. These close aggregate repair counts are consistent with an overall no-harm profile at the task level. Together with Sec~\ref{secr1}, the results show substantive tool-use adjustments alongside broadly retained repair performance.

\begin{figure}
	\centering
	\includegraphics[width=.9\textwidth]{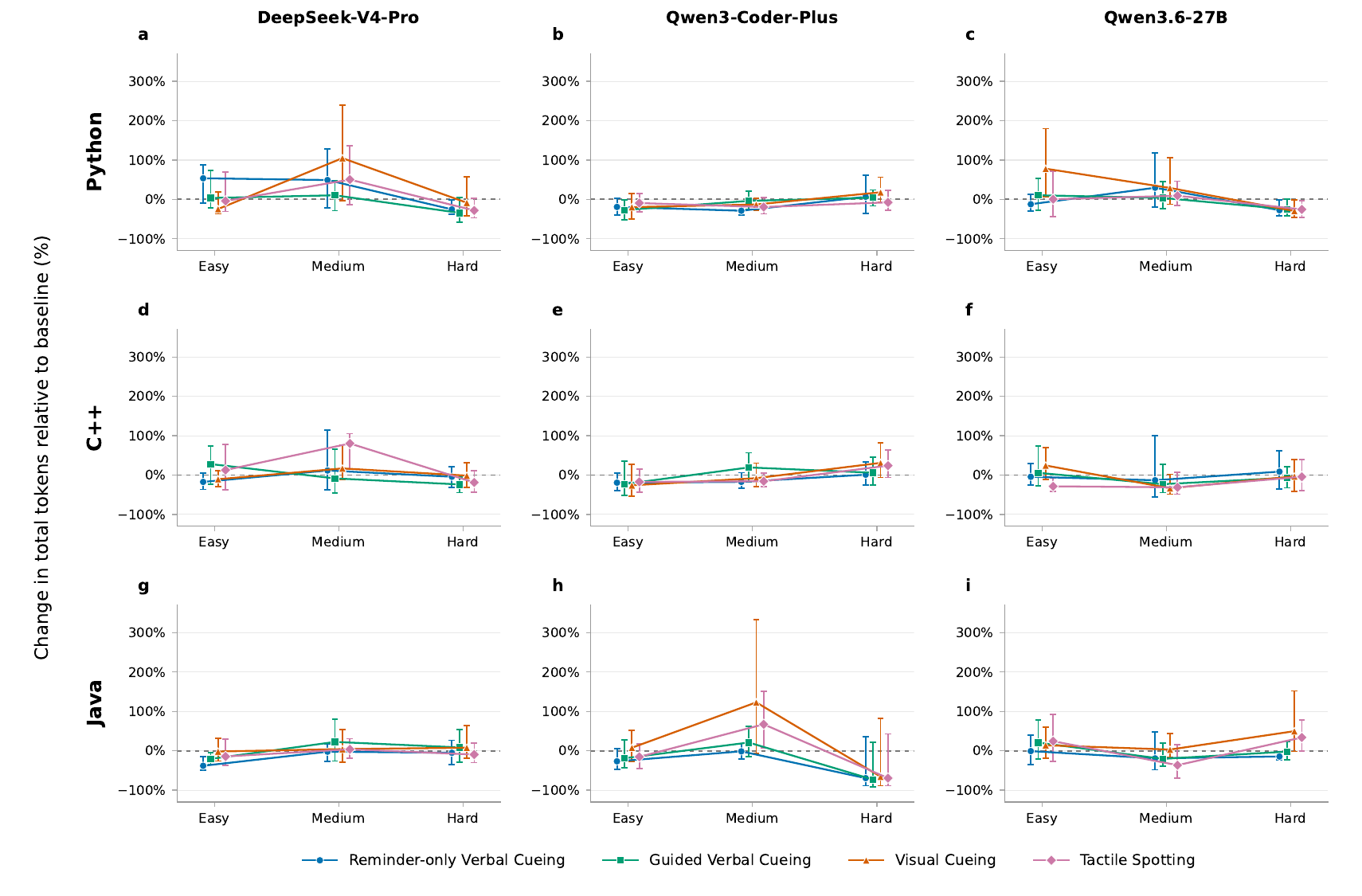}
    \caption{Relative change in total token usage across easy, medium, and hard tasks, grouped by programming language and backbone model. Points show percentage changes relative to the paired baseline, including auxiliary-model tokens where applicable. Lines connect difficulty levels within each assistance condition, and error bars indicate 95\% paired bootstrap confidence intervals.}
    \label{fig:token_difficulty}
\end{figure}

\textbf{Task difficulty further shapes the resource profile of runtime assistance.}
In Figure~\ref{fig:token_difficulty}, all four assistance conditions have lower token-usage point estimates on hard Python instances for DeepSeek-V4-Pro and Qwen3.6-27B, and on hard Java instances for Qwen3-Coder-Plus. These patterns indicate opportunities to shorten execution on difficult tasks alongside small changes in aggregate repair success. The hard C++ estimates for Qwen3-Coder-Plus, however, are near or above baseline, and many intervals include zero. Resource effects therefore depend jointly on task difficulty, the backbone's execution behavior, and the overhead of assistance.

\textbf{The remaining task-level challenge is to support autonomous tool use while preserving successful repairs and avoiding unnecessary execution effort.}
The behavioral findings in Sec~\ref{secr1} and Sec~\ref{secr2} connect runtime assistance with substantive tool-use adjustments, while task-level results reveal individual gains and regressions. For example, reminder-only verbal cueing with Qwen3.6-27B on Java produces four newly passed instances and three regressions, yielding a net gain of one. Lower token totals likewise require examining whether changed trajectories remove unnecessary actions or curtail useful exploration. Further work should investigate intervention points and assistance forms that support tool-use adjustment, retain successful baseline repairs, and reduce avoidable work across backbones.

\begin{rqbox}
\textbf{Answer to RQ3:}

ToolPulse supports agents' adjustment of tool use while broadly retaining observed repair performance, with aggregate task outcomes consistent with an overall no-harm profile. Substantive behavioral adjustments are not accompanied by a corresponding increase in aggregate repair success. Runtime support can also reduce execution expenditure in selected settings, with resource benefits varying across backbone, task, and assistance form. Leveraging assistance that supports autonomous tool use, preserves individual successful repairs, and reduces avoidable work remains an open challenge.
\end{rqbox}

\section{Discussion}\label{sec6}
\subsection{The Role and Value of ToolPulse in Agent Harness}

Developing autonomous agents requires a systems engineering approach that integrates model capabilities, tools, and runtime support into a coherent execution process. An agent harness serves as a programmable engineering tool for organizing, configuring, and extending the interactions among these components. It provides the execution infrastructure connecting model reasoning to tools, feedback, and task state. Designs for long-running agents show how this infrastructure sustains work through environment initialization, progress records, and session handoffs \citep{Young2025Harness}. Within this infrastructure, ToolPulse serves as a runtime support layer for autonomous tool use. It maintains evidence about ongoing interactions and returns selective assistance, giving the harness an active role in helping agents adjust their tool use as tasks unfold.

Table~\ref{tab:harness_comparison} summarizes related approaches and ToolPulse’s corresponding advantages. Its distinctive engineering contribution is the joint treatment of local execution state and temporal interaction patterns at pre- and post-execution checkpoints. This combination connects evidence about individual calls with patterns emerging across interactions and makes both available for targeted assistance. The agent retains responsibility for interpreting the assistance and selecting subsequent actions.

\begin{table}[t]
\centering
\footnotesize
\setlength{\tabcolsep}{4pt}
\renewcommand{\arraystretch}{1.16}
\caption{Related approaches and ToolPulse's corresponding engineering advantages for supporting autonomous tool use.}
\label{tab:harness_comparison}
\begin{tabularx}{\linewidth}{@{}p{0.15\linewidth}X X@{}}
\toprule
\textbf{Related work} & \textbf{Main approach} & \textbf{ToolPulse's advantage} \\
\midrule
ACE \citep{Zhang2026ACE}
& Supplies accumulated playbook strategies as context for new tasks, refining reusable guidance through reflection and curation of execution experience.
& Activates guidance when a tool-use symptom is detected, matching the cue and its delivery point to current execution evidence. \\
\addlinespace
Wink \citep{Nanda2026Wink}
& Inspects trajectories asynchronously at fixed step intervals, providing corrective guidance when an observer’s assessment becomes available to the agent.
& Inspects every tool interaction before and after execution, enabling call interception and feedback-grounded assistance at the corresponding decision point. \\
\addlinespace
LivePlan \citep{liu2026onlinemonitoringcorrectivesteering}
& Detects drift through graph revisits, repeated cycles, and prolonged or missing problem-solving phases, then supplies corresponding corrective advice.
& Combines semantic sequence detection with continuous interaction-feature similarity, adding approximate multistep recurrence to the conditions that can trigger assistance. \\
\addlinespace
Outcome Monitors \citep{Panthi2026OutcomeMonitors}
& Checks individual tool results against outcome contracts and appends violation receipts that identify diagnostic evidence and publicly available recovery tools.
& Extends monitoring to relationships across calls, capturing functional cycles and strategic fixation even when individual tool results remain valid. \\
\addlinespace
Sentry \citep{Ji2026Sentry}
& Uses LLM-based failure diagnosis and an evolving playbook of verified recoveries to select guidance for the agent’s current difficulty.
& Determines intervention triggers through executable state and temporal criteria, enabling direct replay of detection decisions from recorded tool interactions. \\
\bottomrule
\end{tabularx}
\end{table}

\subsection{Reference Implementation and Methodological Implication}

Our empirical findings support ToolPulse's role in helping agents adjust their tool use during autonomous task execution. Extending this support to broader engineering settings requires accommodating variation in agent frameworks, tool interfaces, and execution workflows. We therefore provide a reference implementation of ToolPulse that translates the runtime assistance investigated in this study into reusable components for agent execution environments. It packages state-flow tracking, symptom detection, and assistance generation behind explicit interfaces connected to action proposal, tool execution, and feedback delivery. This organization makes integration concrete: the host supplies action interpretation, task-relevant semantics, and lifecycle connections, while ToolPulse maintains the interaction state and generates assistance.

\begin{figure}
    \centering
    \includegraphics[
        width=\linewidth,
        height=0.70\textheight,
        keepaspectratio
    ]{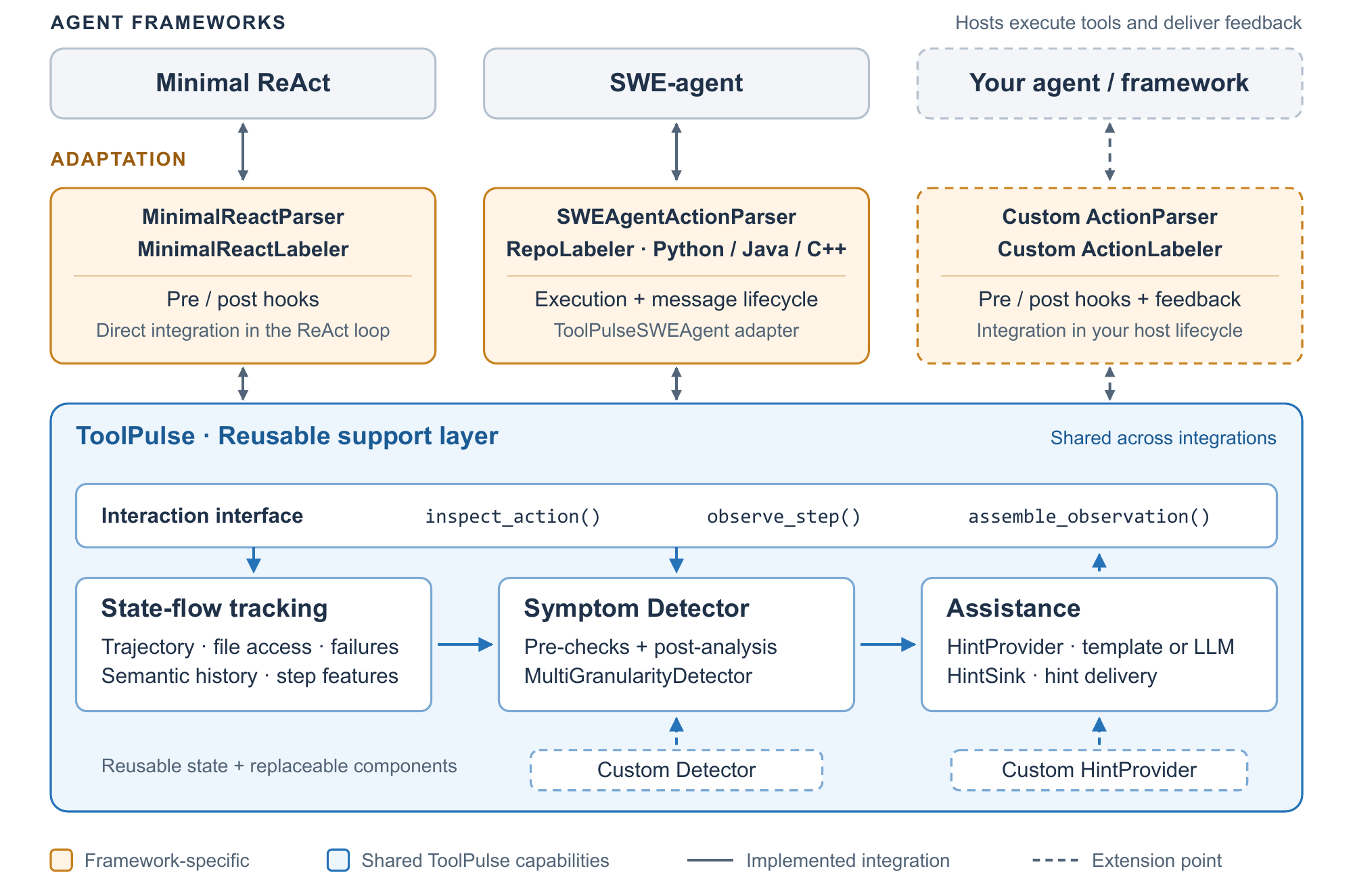}
    \caption{ToolPulse reference implementation. Framework-specific parsers, semantic labelers, and lifecycle connections link agent frameworks to shared state-flow tracking, symptom detection, and assistance components. Minimal ReAct demonstrates explicit integration at individual interaction steps, while the SWE-agent adapter encapsulates integration with an existing framework's execution and message lifecycle. Orange denotes framework-specific adaptation, and blue denotes shared capabilities. Solid connections indicate implemented integrations; dashed elements identify extension points.}
    \label{fig:toolpulse_integration}
\end{figure}

Figure~\ref{fig:toolpulse_integration} presents the component organization. Framework-specific adaptation comprises an action parser, a semantic labeler, and lifecycle integration. The parser translates host actions into a common representation; the labeler supplies functional categories, file-access information, and error signatures. The shared support layer maintains interaction history, observed file state, failure records, semantic sequences, and step features. These records support the \texttt{MultiGranularityDetector}, while \texttt{HintProvider} generates assistance and \texttt{HintSink} handles its delivery. The interaction interface exposes \texttt{inspect\_action()}, \texttt{observe\_step()}, and \texttt{assemble\_observation()} for pre-execution inspection, post-execution analysis, and feedback assembly, respectively.

To demonstrate ToolPulse's integration at different levels of agent execution, we provide two examples: Minimal ReAct and SWE-agent. Minimal ReAct provides step-level integration: ToolPulse is explicitly invoked before tool execution and during post-execution analysis and feedback assembly within an editable interaction loop. Scripted decisions and executable in-memory tools make this sequence reproducible and expose each integration point. SWE-agent provides framework-level integration: a reusable adapter encapsulates the connections to an existing agent's execution and message lifecycle, using repository-oriented semantics for Python, Java, and C++. It handles native tool calls and their response identifiers, execution retries, task initialization, and feedback assembly within the host's output limits.

Motivated by our fine-grained findings on agents' responses across tool-use symptoms and assistance strategies, we organize detection, hint generation, and feedback assembly as separately configurable components. Concise reminders provide a lightweight starting point, reflecting their highest observed aggregate substantive-response rate. Guided templates address concrete corrective requirements, while replaceable feedback providers accommodate the symptom-dependent effects of information organization. Auxiliary-model advice remains separately configurable, reflecting its variable effectiveness relative to simpler cueing. This organization connects assistance choices to the adjustments agents need to make, rather than treating additional guidance or inference as an automatic improvement. Detection thresholds, recurrence rules, and host policies regulate intervention timing and action interception. Documented interfaces and tests support examining these choices through behavioral responses, task outcomes, and resource expenditure.

Users can adopt the supplied integrations, customize symptom-specific templates, or connect another framework through its action parser, semantic vocabulary, and lifecycle hooks. The \texttt{Detector}, \texttt{HintProvider}, and \texttt{HintSink} interfaces support extensions to monitoring, advice generation, and delivery. This organization allows users to reuse the existing state tracking and host connections while concentrating development on new task semantics or assistance mechanisms. Our continued development will focus on progress-aware detection and adaptive intervention policies that use the current execution state, detected symptoms, and prior responses to determine when to intervene, which assistance to provide, and how strongly to intervene.

Research on agentic tooling benefits from examining how agents use feedback, adjust their actions, and sustain task progress alongside final outcomes. This study explores a process-oriented approach: operationalizing tool-use symptoms, introducing targeted assistance, and assessing trigger validity, behavioral cessation, and substantive corrective responses. Connecting these observations with task outcomes provides a basis for investigating how runtime support contributes to autonomous tool use. The reference implementation enables further research through separately configurable detection and assistance mechanisms.

\section{Threats to Validity}\label{sec7}

\textbf{Backbone and auxiliary models.}
The observed effects depend on the backbone's repair capability, tool-calling habits, and responsiveness to external assistance. Task success is constrained by its ability to understand the defect and construct a correct patch, while its characteristic tool-use patterns shape the symptoms encountered and the opportunities for intervention. These properties may vary independently, allowing models with similar repair performance to respond differently to the same assistance. The three evaluated backbones consequently provide distinct conditions for assessing ToolPulse, and the relative value of assistance may vary with other models and execution configurations. Tactile spotting additionally depends on the complementarity between the backbone and auxiliary model, particularly whether the latter supplies useful guidance beyond what the backbone can infer from the existing context.

\textbf{Task selection and execution environment.}
We evaluate a subset of the benchmark selected through stratified random sampling according to estimated repair complexity, incorporating task and reference-patch characteristics. Complexity stratification provides coverage across the estimated difficulty levels, but defect types and repair scopes may still be unevenly represented within those strata. The computational cost of individual executions also makes exhaustive evaluation demanding: across the 30 sampled Python instances, unassisted DeepSeek-V4-Pro runs consume an average of 2,031,739 tokens, with a maximum of 11,952,141. The findings therefore depend on the task composition represented by the sample. Some sampled instances also require resources or platforms unavailable in the experimental environment.\footnote{For example, \texttt{camel-ai\_\_camel-1363} requires accessing documentation supplied through URLs and reviewing components against the updated documentation, whereas the experimental environment has no network access. \texttt{ytdl-org\_\_youtube-dl-32741} requires debugging a Windows executable, while the experiments run on Linux.} Such prerequisites affect the conditions under which task completion and resource expenditure are assessed, alongside the intrinsic repair difficulty.

\textbf{Mission context and prompt formulation.}
The effects of runtime assistance depend on how prompt formulation aligns with the agent's mission context. Mission objectives, operating conditions, and workflows shape the exploration, validation, and recovery required of the agent, influencing both the difficulties it encounters and the adjustments needed to sustain progress. Prompt wording, specificity, and information organization further influence how the agent interprets a cue and incorporates it into ongoing work. The observed response rates and relative effectiveness of assistance forms therefore reflect the evaluated mission context and concrete prompt implementations, and may vary across application settings and alternative formulations of the same assistance.

\textbf{Execution variability.}
Execution paths may vary even when the model, task instance, and assistance strategy remain unchanged. Different assistance conditions introduce further variation by influencing subsequent reasoning, tool selection, and feedback. Paired comparisons preserve task identity, while the resulting trajectories may diverge substantially. Observed differences in patch validation and token usage therefore reflect both assistance effects and execution variability, particularly when pass counts differ by only one or two instances. The paired bootstrap intervals for token usage quantify uncertainty in comparisons of the recorded runs; repeated executions of the same configuration may introduce additional variation.

\section{Conclusion}\label{sec8}
We present ToolPulse, an online middleware that supports autonomous tool use by tracking evolving execution states and interaction patterns and providing reminders and guidance through cueing and spotting for clumsy tool-use behaviors and tool-use detours. Our experiments on long-horizon bug-fix tasks show substantive corrective responses to targeted assistance and provide evidence consistent with an overall no-harm profile at the task level. Concise reminders provide effective support, whereas more elaborate assistance offers symptom-dependent benefits and can even be less effective than concise reminders. Auxiliary-model advice provides no consistent advantage over simpler cueing. Resource savings are observed in selected settings and depend on the combination of backbone, task, and assistance form.

These findings motivate a broader engineering role for the agent's runtime environment. An environment can contribute to task progress by maintaining evidence about how tool use unfolds and returning assistance when that evidence warrants attention. ToolPulse provides an initial realization of this role, with the executing agent retaining responsibility for subsequent decisions. This perspective makes supporting agents in adjusting their tool use and sustaining autonomous task progress an explicit design objective alongside task-solving capability. It also places the quality of assistance in the relationship between execution evidence, the agent's response, and the resulting task outcome.

A promising direction is to develop runtime support that adapts to the working rhythms of different agents throughout their agentic loops. Such support would need to determine when different levels of intervention help the agent resume productive task execution. ToolPulse offers a starting point for investigating whether these capabilities can become reusable infrastructure across agents and task settings. The longer-term opportunity is to develop an execution infrastructure in which agents, tools, and runtime support work together to sustain reliable autonomy across complex tasks and changing environments.
\appendix
\section{Reproducibility Details of the Pilot Study}\label{secA1}
\subsection{Tasks and Run Configuration}
\paragraph{Task instances.}
The experiment of pilot study uses 30 task instances from the Python bug-fix subset of OmniCode, with 10 instances from each of the easy, medium, and hard sampling levels. These levels reflect the patch-based complexity proxy described in the main text. Table~\ref{tab:app_pilot_tasks} lists the instance IDs that define the sample. You can find the benchmark release in this \href{https://github.com/seal-research/OmniCode}{GitHub Repository}. The sampling seed is 20260609.

\begin{table}[pos=t]
\centering
\small
\caption{The 30 Python task instances used in the pilot study.}
\label{tab:app_pilot_tasks}
\begin{tabular}{rlp{0.68\linewidth}}
\toprule
\textbf{No.} & \textbf{Difficulty} & \textbf{Task instance ID} \\
\midrule
1 & Easy & astropy\_\_astropy-7166 \\
2 & Easy & django\_\_django-13516 \\
3 & Easy & django\_\_django-13794 \\
4 & Easy & fastapi\_\_fastapi-1524 \\
5 & Easy & psf\_\_requests-1921 \\
6 & Easy & pytest-dev\_\_pytest-7982 \\
7 & Easy & scrapy\_\_scrapy-5526 \\
8 & Easy & sympy\_\_sympy-15809 \\
9 & Easy & sympy\_\_sympy-23262 \\
10 & Easy & ytdl-org\_\_youtube-dl-23199 \\
\addlinespace
11 & Medium & django\_\_django-11749 \\
12 & Medium & django\_\_django-16493 \\
13 & Medium & matplotlib\_\_matplotlib-22865 \\
14 & Medium & matplotlib\_\_matplotlib-25332 \\
15 & Medium & pytest-dev\_\_pytest-10051 \\
16 & Medium & pytest-dev\_\_pytest-7432 \\
17 & Medium & scikit-learn\_\_scikit-learn-26194 \\
18 & Medium & sphinx-doc\_\_sphinx-10449 \\
19 & Medium & sympy\_\_sympy-14531 \\
20 & Medium & sympy\_\_sympy-24661 \\
\addlinespace
21 & Hard & camel-ai\_\_camel-1363 \\
22 & Hard & camel-ai\_\_camel-1627 \\
23 & Hard & django\_\_django-11333 \\
24 & Hard & django\_\_django-13820 \\
25 & Hard & keras-team\_\_keras-20380 \\
26 & Hard & mwaskom\_\_seaborn-3187 \\
27 & Hard & scrapy\_\_scrapy-5833 \\
28 & Hard & sphinx-doc\_\_sphinx-10673 \\
29 & Hard & sphinx-doc\_\_sphinx-11510 \\
30 & Hard & ytdl-org\_\_youtube-dl-32741 \\
\bottomrule
\end{tabular}
\end{table}

\paragraph{Experimental configurations.}
We evaluate SWE-agent with DeepSeek-V4-Pro and Qwen3-Coder-Plus under four configurations: Baseline, Sequitur, Semantic Sequitur, and Matrix Profile. The temperature of each model is 0, and Top-p is 1. You can find the agent framework in this \href{https://github.com/seal-research/SWE-agent/tree/04bb72b888088b966f5539153b15b20e0218d348}{GitHub Repository}. Each model--task--configuration combination has one run, yielding $2\times30\times4=240$ runs. Comparisons pair runs by task instance ID within the same backbone model. Baseline uses no additional intervention prompts; the three intervention configurations use the observation-level prompting procedure described below.

\subsection{Detection Rules and Intervention Prompts}
\paragraph{Interaction timing.}
We distinguish a \emph{detection trigger} from an \emph{injected reminder}. A trigger satisfies a detection rule, whereas a reminder is a message that passes the online suppression policy and is actually delivered to the agent. If multiple Action types qualify at the same step, include all of them.

\paragraph{Shared Action conditions.}
Duplicate Action occurs when two consecutive calls have exactly the same tool name and arguments. The pilot therefore uses a threshold of two consecutive identical calls. Argument equality follows raw-string or canonicalized comparison, including whitespace and key-order handling.

Failed Action occurs when a tool call returns an execution error. The executable rule is tool-specific parsing rules. The treatment of expected test failures, intentional error reproduction, and infrastructure errors have exact exclusion rules. Tool names and error types are normalized using normalization rules and error-type taxonomy.

Loop Action and Custom Tool Candidate depend on the trajectory representation. Their strategy-specific conditions are given below. For Custom Tool Candidate, the additional eligibility conditions are minimum recurrence and candidate-selection filters. The candidate-generation rule is separate from whether the agent subsequently creates or invokes a custom tool.

\textit{Sequitur.}
We encode each tool name as one discrete symbol and apply Sequitur to the resulting sequence. A recurring grammar rule (with a minimum expansion length of 3) qualifies as Loop Action under minimum 3 occurrence count. The rule for selecting a recurring sequence as a Custom Tool Candidate requires a minimum grammar expansion length of 3 and 4 non-consecutive repetitions.

\textit{Semantic Sequitur.}
We assign one deterministic label to each interaction and apply Sequitur to the label sequence. The label alphabet is
$\{\mathrm{Ls},\mathrm{Lt},\mathrm{Lb},\mathrm{E},\mathrm{P},
\mathrm{Ps},\mathrm{Pi},\mathrm{Pr},\mathrm{Vp},\mathrm{Vf},
\mathrm{Ve},\mathrm{G},\mathrm{Emp}\}$,
with the meanings given in the main text. The Sequitur implementation and the Loop Action and Custom Tool Candidate rules are the same as Sequitur group but using semantic label instead of raw tool.

\textit{Matrix Profile.}
For vector handling, we encode each action into a 36-dimensional vector laid out as the z-scored thought length (1 dim), a triple-hash block for the tool name (16 dims), a triple-hash block for the serialized args (16 dims), plus observation length, execution time, and an error bit (1 dim each). Categorical features use the hashing trick: the tool name and JSON-serialized args are each mapped by md5, blake2b, and sha256 into three distinct dimensions of a 16-dim block (linear probing on collision), set to 1, L2-normalized, and scaled by weights (tool 0.80, args 0.45), so tool identity and argument shape share one continuous space with no vocabulary to maintain; scalar features pass through log1p and are then z-scored with mean and variance estimated from all history except the newest step, which prevents an outlier from diluting its own distance.

For window updating, when each action comes, first drops the oldest entries beyond 500 and clears the motif adjacency graph; then computes, for each window size (3/5/8), the matrix profile of only the newest window by measuring its distance to every earlier non-overlapping window — each window stacks into a (size, 36) matrix and the distance is the Frobenius norm of the difference divided by sqrt(size), a per-step RMS distance that keeps thresholds comparable across sizes — storing the minimum as the profile value while adding bidirectional adjacency edges to any earlier window within distance 1.25 with tool-sequence similarity greater than or equal to 0.5. Both downstream detectors build on these incremental artifacts: loop detection checks whether the max pairwise distance among the last three consecutive same-size windows stays less than or equal to 1.10, and macro detection expands a connected component from the current window in the adjacency graph and selects non-overlapping windows into a cluster.

\paragraph{Intervention templates.}
The following blocks give the complete text of the four prompt templates. Braced fields are filled from the current trajectory.

\textbf{Duplicate Action prompt.}
\\For empty actions:
\begin{PromptBlock}
I notice an empty action was produced twice in a row. Make sure the next response contains an executable action.
\end{PromptBlock}
For normal actions:
\begin{PromptBlock}
I notice the exact same action was used twice in a row: {tool} {args}. Read the latest observation carefully before casting that spell again.
\end{PromptBlock}

\textbf{Loop Action prompt.}
\\For Sequitur and Semantic Sequitur:
\begin{PromptBlock}
I sense a repeating rhythm, seen {repeat_times} times: {pattern_text}. This may be a loop; you may consider breaking the pattern with a different strategy.
\end{PromptBlock}
For Matrix Profile:
\begin{PromptBlock}
I notice {repeat_times} consecutive windows with approximately similar behavior. A representative tool pattern is: {pattern_text}. The actions are not necessarily identical, but this may indicate repeated investigation, editing, or validation without enough progress. Review the latest results and consider a materially different next step.
\end{PromptBlock}

\textbf{Failed Action prompt.}
\begin{PromptBlock}
I sense a {error_type} error from the previous {tool} action. Before your next move, think through what caused the error.
\end{PromptBlock}

\textbf{Custom Tool Candidate prompt.}
\begin{PromptBlock}
I notice {workflow_text} recurring around this representative pattern: {pattern_text}. The individual actions and arguments may differ, so first verify that the workflow is stable and reusable. If it is, consider a custom tool such as `{suggested_name}`.
\end{PromptBlock}

the workflow\_text above varies for patterns, including approximately similar workflows, similar edit-and-validation workflows and similar targeted search-and-read workflows.

\section{Reproducibility Details of the Main Experiments}\label{secA2}
\subsection{Tasks and Run Configuration}
\paragraph{Task instances.}
The main experiment uses 90 bug-fix instances from OmniCode, comprising 30 instances each in Python, C++, and Java. The benchmark source and the complexity proxy used to define the easy, medium, and hard sampling levels follow the pilot study described in Appendix~\ref{secA1}. Tables~\ref{tab:app_pilot_tasks}, \ref{tab:app_main_cpp_tasks}, and \ref{tab:app_main_java_tasks} lists the instance IDs that define the sample. All experimental conditions use the same task instances within each backbone model. The sampling seed is 20260609.

\begin{table}[pos=t]
\centering
\small
\caption{The 30 C++ task instances used in the main experiments.}
\label{tab:app_main_cpp_tasks}
\begin{tabular}{rlp{0.68\linewidth}}
\toprule
\textbf{No.} & \textbf{Difficulty} & \textbf{Task instance ID} \\
\midrule
1 & Easy & catchorg\_\_Catch2\_2288 \\
2 & Easy & fmtlib\_\_fmt\_1390 \\
3 & Easy & fmtlib\_\_fmt\_1407 \\
4 & Easy & fmtlib\_\_fmt\_2070 \\
5 & Easy & fmtlib\_\_fmt\_2310 \\
6 & Easy & fmtlib\_\_fmt\_2940 \\
7 & Easy & fmtlib\_\_fmt\_3713 \\
8 & Easy & fmtlib\_\_fmt\_3729 \\
9 & Easy & nlohmann\_\_json\_18 \\
10 & Easy & nlohmann\_\_json\_2989 \\
\addlinespace
11 & Medium & catchorg\_\_Catch2\_1448 \\
12 & Medium & catchorg\_\_Catch2\_1609 \\
13 & Medium & catchorg\_\_Catch2\_1614 \\
14 & Medium & catchorg\_\_Catch2\_1616 \\
15 & Medium & nlohmann\_\_json\_1555 \\
16 & Medium & nlohmann\_\_json\_3446 \\
17 & Medium & nlohmann\_\_json\_3564 \\
18 & Medium & nlohmann\_\_json\_944 \\
19 & Medium & simdjson\_\_simdjson\_2016 \\
20 & Medium & simdjson\_\_simdjson\_2026 \\
\addlinespace
21 & Hard & catchorg\_\_Catch2\_1608 \\
22 & Hard & catchorg\_\_Catch2\_2128 \\
23 & Hard & catchorg\_\_Catch2\_2187 \\
24 & Hard & catchorg\_\_Catch2\_2394 \\
25 & Hard & fmtlib\_\_fmt\_1171 \\
26 & Hard & fmtlib\_\_fmt\_4286 \\
27 & Hard & nlohmann\_\_json\_1221 \\
28 & Hard & nlohmann\_\_json\_2332 \\
29 & Hard & simdjson\_\_simdjson\_1896 \\
30 & Hard & simdjson\_\_simdjson\_2178 \\
\bottomrule
\end{tabular}
\end{table}

\begin{table}[pos=t]
\centering
\small
\caption{The 30 Java task instances used in the main experiments.}
\label{tab:app_main_java_tasks}
\begin{tabular}{rlp{0.68\linewidth}}
\toprule
\textbf{No.} & \textbf{Difficulty} & \textbf{Task instance ID} \\
\midrule
1 & Easy & alibaba\_\_fastjson2\_2559 \\
2 & Easy & apache\_\_dubbo\_11781 \\
3 & Easy & elastic\_\_logstash\_15697 \\
4 & Easy & fasterxml\_\_jackson-core\_1142 \\
5 & Easy & fasterxml\_\_jackson-core\_1208 \\
6 & Easy & fasterxml\_\_jackson-databind\_4050 \\
7 & Easy & fasterxml\_\_jackson-dataformat-xml\_544 \\
8 & Easy & fasterxml\_\_jackson-dataformat-xml\_590 \\
9 & Easy & google\_\_guava\_2945 \\
10 & Easy & googlecontainertools\_\_jib\_2688 \\
\addlinespace
11 & Medium & elastic\_\_logstash\_13902 \\
12 & Medium & elastic\_\_logstash\_14571 \\
13 & Medium & elastic\_\_logstash\_14897 \\
14 & Medium & elastic\_\_logstash\_16094 \\
15 & Medium & elastic\_\_logstash\_16579 \\
16 & Medium & fasterxml\_\_jackson-databind\_4087 \\
17 & Medium & fasterxml\_\_jackson-dataformat-xml\_531 \\
18 & Medium & fasterxml\_\_jackson-dataformat-xml\_644 \\
19 & Medium & google\_\_gson\_1787 \\
20 & Medium & mockito\_\_mockito\_3133 \\
\addlinespace
21 & Hard & elastic\_\_logstash\_15928 \\
22 & Hard & elastic\_\_logstash\_16681 \\
23 & Hard & fasterxml\_\_jackson-core\_922 \\
24 & Hard & fasterxml\_\_jackson-databind\_4132 \\
25 & Hard & fasterxml\_\_jackson-databind\_4320 \\
26 & Hard & fasterxml\_\_jackson-databind\_4426 \\
27 & Hard & google\_\_gson\_1391 \\
28 & Hard & google\_\_guava\_6586 \\
29 & Hard & googlecontainertools\_\_jib\_4144 \\
30 & Hard & mockito\_\_mockito\_3167 \\
\bottomrule
\end{tabular}
\end{table}

\paragraph{Experimental configurations.}
We evaluate SWE-agent with DeepSeek-V4-Pro, Qwen3-Coder-Plus, and Qwen3.6-27B under five configurations: Baseline, reminder-only verbal cueing, guided verbal cueing, visual cueing, and tactile spotting. The SWE-agent version and backbone sampling parameters are the same as in the pilot study: temperature is 0 and top-p is 1. Each model--task--configuration combination has one run, yielding $3\times90\times5=1{,}350$ runs, including 270 baseline runs and 1,080 assisted runs. Comparisons pair each assisted run with the baseline run for the same task instance and backbone model. Baseline receives no ToolPulse assistance. The four assisted configurations share the same symptom-detection settings and differ in the assistance delivered to the agent. Tactile Spotting uses Qwen3.7-Flash~\citep{qwen37} as the auxiliary model, with temperature set to 0 and top-p to 1.

\subsection{Symptom Detection and Assistance Conditions}

\paragraph{Detection configuration.}
Calling Repetition and Blind Command follow the definitions in the main text. All four assistance strategies share these detection settings. Strategic Fixation uses the Matrix Profile loop-detection implementation and settings described in Appendix~\ref{secA1}. Functional Cycle follows the Sequitur loop-detection implementation and settings in the same appendix, applied to the semantic action labels in Table~\ref{tab:bugfix_labels}. Failure Inertia uses a lookback window of 8 steps.

\paragraph{Assistance templates.}
The following blocks provide the templates used under the four assistance strategies.

\textbf{Reminder-only verbal cueing templates.}

\textit{Calling Repetition.}
\\For empty actions:
\begin{PromptBlock}
Tool-use notice: You invoked an empty action consecutively for 3 times.
\end{PromptBlock}
For normal actions:
\begin{PromptBlock}
Tool-use notice: You called {tool} with exactly the same arguments in 3 consecutive steps.
\end{PromptBlock}

\textit{Functional Cycle.}
\begin{PromptBlock}
Tool-use notice: You repeated the action pattern {pattern_text} for {repeat_times} times.
\end{PromptBlock}

\textit{Strategic Fixation.}
\begin{PromptBlock}
Tool-use notice: You took highly similar tool-use behavior across consecutive steps, each spanning {production} interactions.
\end{PromptBlock}

\textit{Blind Command.}
\begin{PromptBlock}
Tool-use notice: You tried to edit {path} using {tool} before inspecting the file in the current task.
\end{PromptBlock}

\textit{Failure Inertia.}
\begin{PromptBlock}
Tool-use notice: You received {error_type} error again from {tool} without taking an effective recovery action after the previous error.
\end{PromptBlock}

\textbf{Guided verbal cueing templates.}

\textit{Calling Repetition.}
\\For empty actions:
\begin{PromptBlock}
Tool-use notice: You invoked an empty action consecutively for 3 times.
Guidance: Examine the previous result before retrying, and make a materially different call only when new evidence justifies it.
\end{PromptBlock}
For normal actions:
\begin{PromptBlock}
Tool-use notice: You called {tool} with exactly the same arguments in 3 consecutive steps.
Guidance: Examine the previous result before retrying, and make a materially different call only when new evidence justifies it.
\end{PromptBlock}

\textit{Functional Cycle.}
\begin{PromptBlock}
Tool-use notice: You repeated the action pattern {pattern_text} for {repeat_times} times.
Guidance: Consider whether this repetition is still contributing to progress. If not, reconsider the current tool-use strategy.
\end{PromptBlock}

\textit{Strategic Fixation.}
\begin{PromptBlock}
Tool-use notice: You took highly similar tool-use behavior across consecutive steps, each spanning {production} interactions.
Guidance: Assess whether your current work has stalled or delivered only limited progress on the task. If it is, revise your strategy if progress is insufficient.
\end{PromptBlock}

\textit{Blind Command.}
\begin{PromptBlock}
Tool-use notice: You tried to edit {path} using {tool} before inspecting the file in the current task.
Guidance: Inspect the target file and its relevant context before making an edit grounded in the observed content.
\end{PromptBlock}

\textit{Failure Inertia.}
\begin{PromptBlock}
Tool-use notice: You received {error_type} error again from {tool} without taking an effective recovery action after the previous error.
Guidance: Determine whether this error reflects an actual failure or an expected outcome of the current action. If it is an actual failure, examine its cause and address any unmet preconditions before proceeding.
\end{PromptBlock}

\textbf{Visual cueing templates.}

\textit{Calling Repetition.}
\\For empty actions:
\begin{PromptBlock}
{
  "observation": original observation information,
  "hint": "Tool-use notice: You invoked an empty action consecutively for 3 times."
}
\end{PromptBlock}
For normal actions:
\begin{PromptBlock}
{
  "observation": original observation information,
  "hint": "Tool-use notice: You called {tool} with exactly the same arguments in 3 consecutive steps."
}
\end{PromptBlock}

\textit{Functional Cycle.}
\begin{PromptBlock}
{
  "observation": original observation information,
  "hint": "Tool-use notice: You repeated the action pattern {pattern_text} for {repeat_times} times."
}
\end{PromptBlock}

\textit{Strategic Fixation.}
\begin{PromptBlock}
{
  "observation": original observation information,
  "hint": "Tool-use notice: You took highly similar tool-use behavior across consecutive steps, each spanning {production} interactions."
}
\end{PromptBlock}

\textit{Blind Command.}
\begin{PromptBlock}
{
  "observation": original observation information,
  "hint": "Tool-use notice: You tried to edit {path} using {tool} before inspecting the file in the current task."
}
\end{PromptBlock}

\textit{Failure Inertia.}
\begin{PromptBlock}
{
  "observation": original observation information,
  "hint": "Tool-use notice: You received {error_type} error again from {tool} without taking an effective recovery action after the previous error."
}
\end{PromptBlock}

\textbf{Tactile spotting: prompt for the auxiliary model.}
The following prompts instruct the auxiliary model to generate context-specific assistance for the agent.

\begin{PromptBlock}
You are an experienced software developer. Help examine the current progress and recent tool interactions. When a potentially problematic tool-use behavior is observed, use your experience to assess the situation, perform a focused analysis that reduces the remaining diagnostic burden, and recommend next action.

Software engineering task:
{{TASK_DESCRIPTION}}
      
Observed tool-use behavior:
{{BEHAVIOR_DESCRIPTION}}

Relevant behavior details:
{{METADATA}}

Recent work trajectory:
{{RECENT_TRAJECTORY}}

Provide concise and context-aware assistance that:
1. Examines the recent trajectory and describes what happened in the observed tool-use behavior.
2. Assess whether the behavior is appropriate for the current situation or is hindering progress toward the task.
3. Recommends a concrete next action that advances the software engineering task. Do not merely give generic advice such as "try another approach" or "inspect the error."
4. Do not assume that every flagged behavior is harmful. First determine whether the behavior is intentional or still contributing to progress.
5. Do not invent tool results, code facts, errors, files, or task conditions that are absent from the provided information.
6. Uses direct, concise, and actionable natural language.

Use exactly the following output format:
Tool-use notice: <describe the observed behavior and its relevance to the current progress>
Recommended action: <provide focused assistance and state the concrete next action>
\end{PromptBlock}

RECENT\_TRAJECTORY contains the previous 3 steps of the current step. The content of BEHAVIOR\_DESCRIPTION is as follows:

\textit{Calling Repetition.}
\begin{PromptBlock}
The agent called {{TOOL}} with exactly the same arguments in 3 consecutive steps.
\end{PromptBlock}

\textit{Functional Cycle.}
\begin{PromptBlock}
The agent repeated the action pattern {{PATTERN}} for {{REPEAT_TIMES}} times.
\end{PromptBlock}

\textit{Strategic Fixation.}
\begin{PromptBlock}
The agent took highly similar tool-use behavior across consecutive steps, each spanning {{SIZE}} interactions.
\end{PromptBlock}

\textit{Blind Command.}
\begin{PromptBlock}
The agent tried to edit {{PATH}} using {{TOOL}} before inspecting the file in the current task.
\end{PromptBlock}

\textit{Failure Inertia.}
\begin{PromptBlock}
The agent received {{ERROR_TYPE}} error again from {{TOOL}} without taking an effective recovery action after the previous error.
\end{PromptBlock}

\subsection{Symptom Validation and Response Assessment}

\paragraph{LLM-judge configuration.}
All LLM-based assessments, including the offline assessment of Wandering Flood, use Kimi-K3, with temperature set to 0 and top-p to 1. The assessment context includes up to eight preceding and up to 3 steps following the assessed trigger or candidate step, as specified in the main text. The corresponding prompt templates are provided below.

\textit{Online symptom assessment.}

For Functional Cycle:
\begin{PromptBlock}
You are an expert evaluator specializing in LLM-based coding agents,
and software engineering.

Your task is to evaluate coding agent's tool-use behavior in a part of the trajectory, exactly whether the agent was repeatedly executing a semantically equivalent ordered sequence of actions around the marked trigger step, and whether it made a substantive correction afterward.

TASK DESCRIPTION:
{{TASK_DESCRIPTION}}

HINT SHOWN TO THE AGENT:
{{HINT}}

PRE TRIGGER STEPS:
{{PRE_TRIGGER_STEPS}}

TRIGGER STEP:
{{TRIGGER_STEP}}

POST TRIGGER STEPS:
{{POST_TRIGGER_STEPS}}

Evaluation criteria:

1. Determine whether the pre-trigger trajectory contains repeated,
semantically equivalent action sequences that return to substantially the
same state or objective without obtaining information.
2. Determine whether the agent breaks this pattern after the trigger.

A response is a substantive correction when the agent takes a task-relevant
action that:
- responds to evidence accumulated during previous repetitions;
- changes the underlying approach, target, hypothesis, or information source;
  or
- uses previous observations to make concrete progress.

The following do not constitute a substantive correction:
- changing only wording, formatting, tool aliases, or inconsequential arguments;
- replacing a tool with another tool that serves the same role in the sequence;
- beginning another semantically equivalent instance of the repeated sequence;
- switching to an unrelated action without a task-relevant reason.

General rules:
- The hint is contextual information, not evidence that the behavior occurred
  or was corrected.
- Evaluate semantic intent and trajectory effects rather than surface differences.
- Do not require the overall task to be solved within the provided window.
- If the trajectory does not provide enough evidence, return "unclear".
- Base the judgment only on the supplied information.
- Return valid JSON only, without Markdown or additional text.

Return the following structure:
{
  "trigger_validity": "valid | invalid | unclear",
  "behavior_after_trigger": "persists | ceases | unclear",
  "corrective_response": "substantive | non_substantive | unclear",
  "evidence_step_ids": ["relevant step identifiers"],
  "brief_reason": "A concise explanation of at most 60 words."
}
\end{PromptBlock}

For Strategic Fixation:
\begin{PromptBlock}
You are an expert evaluator specializing in LLM-based coding agents,
and software engineering.

Your task is to evaluate coding agent's tool-use behavior in a part of the trajectory, exactly whether the agent remained committed to the same
ineffective problem-solving strategy around the marked trigger step, and
whether it made a substantive strategic adjustment afterward.

TASK DESCRIPTION:
{{TASK_DESCRIPTION}}

HINT SHOWN TO THE AGENT:
{{HINT}}

PRE TRIGGER STEPS:
{{PRE_TRIGGER_STEPS}}

TRIGGER STEP:
{{TRIGGER_STEP}}

POST TRIGGER STEPS:
{{POST_TRIGGER_STEPS}}

Evaluation criteria:

1. Determine whether the pre-trigger trajectory shows that the agent repeatedly
pursued substantially the same strategy despite obtaining little progress or
evidence that the strategy was inadequate.
2. Determine whether the agent substantively adjusts its strategy after the
trigger.

A response is a substantive correction when the agent:
- revises its working hypothesis or plan based on previous observations;
- changes the relevant target, evidence source, diagnostic method, or solution
  approach; or
- performs a diagnostic action that directly investigates why the previous
  approach stalled.

Continuing to use the same tool is not necessarily evidence of continued
stagnation if the purpose of the tool call or its role in the investigation
has materially changed.

The following do not constitute a substantive correction:
- paraphrasing the same reasoning;
- making minor argument or target variations under the same ineffective strategy;
- switching tools while preserving the same ineffective intent;
- taking different actions that neither address the source of stagnation nor
obtain relevant new evidence.

General rules:
- The hint is contextual information, not evidence that the behavior occurred
  or was corrected.
- Evaluate the underlying strategy rather than superficial action diversity.
- Do not require the overall task to be solved within the provided window.
- If the trajectory does not provide enough evidence, return "unclear".
- Base the judgment only on the supplied information.
- Return valid JSON only, without Markdown or additional text.

Return the following structure:
{
  "trigger_validity": "valid | invalid | unclear",
  "behavior_after_trigger": "persists | ceases | unclear",
  "corrective_response": "substantive | non_substantive | unclear",
  "evidence_step_ids": ["relevant step identifiers"],
  "brief_reason": "A concise explanation of at most 60 words."
}
\end{PromptBlock}

For Failure Inertia:
\begin{PromptBlock}
You are an expert evaluator specializing in LLM-based coding agents,
and software engineering.

Your task is to evaluate coding agent's tool-use behavior in a part of the trajectory, exactly whether the agent repeatedly encountered substantially the same tool failure without taking a meaningful recovery
action, and whether it made a substantive recovery attempt after the
marked trigger step.

TASK DESCRIPTION:
{{TASK_DESCRIPTION}}

HINT SHOWN TO THE AGENT:
{{HINT}}

PRE TRIGGER STEPS:
{{PRE_TRIGGER_STEPS}}

TRIGGER STEP:
{{TRIGGER_STEP}}

POST TRIGGER STEPS:
{{POST_TRIGGER_STEPS}}

Evaluation criteria:

1. Determine whether the pre-trigger trajectory contains repeated tool failures
with substantially the same tool, error type, and underlying failure signature,
without an intervening action that meaningfully addresses the failure.
2. Determine whether the agent takes a substantive recovery action after the
trigger.

A response is a substantive correction when the agent:
- investigates the error or checks a relevant prerequisite;
- changes an argument, command, target, configuration, or environment state in
  a way that addresses the failure;
- repairs the condition responsible for the failure; or
- adopts an alternative tool or strategy that avoids the same failure mode.

The recovery attempt does not need to succeed within the provided window.

The following do not constitute a substantive correction:
- repeating substantially the same failed action;
- merely restating or paraphrasing the error;
- making a cosmetic modification unrelated to the cause of the failure;
- taking an unrelated action that does not contribute to diagnosis or recovery.

General rules:
- The hint is contextual information, not evidence that the behavior occurred
  or was corrected.
- Evaluate whether the response addresses the failure, not merely whether the
  next action is different.
- Do not require the overall task to be solved within the provided window.
- If the trajectory does not provide enough evidence, return "unclear".
- Base the judgment only on the supplied information.
- Return valid JSON only, without Markdown or additional text.

Return the following structure:
{
  "trigger_validity": "valid | invalid | unclear",
  "behavior_after_trigger": "persists | ceases | unclear",
  "corrective_response": "substantive | non_substantive | unclear",
  "evidence_step_ids": ["relevant step identifiers"],
  "brief_reason": "A concise explanation of at most 60 words."
}
\end{PromptBlock}

\textit{Offline Wandering Flood assessment.}
\begin{PromptBlock}
You are an expert evaluator of software engineering agents and coding agent's tool use.

Evaluate the action marked as TRIGGER STEP. It was selected as a candidate because it produced a relatively large observation, took a relatively long time to execute, or both. These signals alone do not mean that the action was inappropriate.

Determine whether the target was an unnecessarily broad or costly exploratory action.

TASK DESCRIPTION
{{TASK_DESCRIPTION}}

PRE TRIGGER STEPS:
{{PRE_TRIGGER_STEPS}}

TRIGGER STEP:
{{TRIGGER_STEP}}

POST TRIGGER STEPS:
{{POST_TRIGGER_STEPS}}

Return yes only if:
1. The action was primarily exploratory or diagnostic.
2. Given what the agent already knew, its scope was substantially broader than necessary and a clearly more focused alternative was available.
3. The high cost was meaningfully caused by the agent's tool or argument choices, rather than unavoidable tool overhead or unexpected latency.
4. The result provided little task-relevant new information relative to its cost and did not materially support the following steps.

Return no if the action was reasonably targeted, necessary under the available information, produced useful evidence, was not exploratory, or became expensive for reasons the agent could not reasonably anticipate.

Return uncertain if deciding whether the action was justified or useful would require information outside the supplied local trajectory. Judge the action based on information available before it was executed; use following steps only to assess its information value. Do not infer missing history.

Return the following structure:
{
  "verdict": "yes | no | uncertain",
  "information_gain": "sufficient | limited | unclear",
  "rationale": "A concise evidence-based explanation."
}
\end{PromptBlock}

\paragraph{Manual verification.}
\textit{Online symptom assessment.}
One expert randomly inspected 100 LLM-judge outputs, checking the categorical judgments and their supporting evidence. Agreement with the expert was 87\% for \texttt{trigger\_validity}, 97\% for \texttt{behavior\_after\_trigger}, and 98\% for \texttt{corrective\_response}. In all 100 outputs, the cited step IDs referred to existing trajectory steps, and the expert judged those steps to support the corresponding \texttt{brief\_reason}. Table~\ref{tab:judge_manual_verification} reports the agreement counts and expert revisions.

\textit{Wandering Flood assessment.}
Expert review of 100 Wandering Flood judgments yields 90\% agreement on \texttt{verdict}. All 20 positive judgments are confirmed. Of the 80 negative judgments, 70 are confirmed (87.5\%) and 10 are revised to positive; no uncertain cases are recorded. Separately, to examine potential cases missed by the first-stage screen, we assess sampled Retrieve steps that did not pass the screen but fell within the 90th--95th percentile band, following the procedure described in the main text. The LLM judge labels 129 of these 1,470 steps as Wandering Flood (8.8\%). Given the role of the first stage as a low-cost candidate screen, we consider this proportion acceptable for the scope of the present study.

\begin{table}[t]
\centering
\small
\caption{Manual verification of 100 randomly sampled LLM-judge outputs by one expert. Arrows indicate revisions from the judge's label to the expert's label.}
\label{tab:judge_manual_verification}
\resizebox{\linewidth}{!}{%
\begin{tabular}{lcp{0.48\linewidth}}
\toprule
\textbf{Assessment field}
& \textbf{Agreement}
& \textbf{Expert revisions or evidence check} \\
\midrule

\texttt{trigger\_validity}
& 87/100 (87\%)
& 12 \texttt{valid}$\rightarrow$\texttt{invalid};
  1 \texttt{unclear}$\rightarrow$\texttt{invalid}. \\

\addlinespace
\texttt{behavior\_after\_trigger}
& 97/100 (97\%)
& 2 \texttt{ceases}$\rightarrow$\texttt{persists};
  1 \texttt{persists}$\rightarrow$\texttt{unclear}. \\

\addlinespace
\texttt{corrective\_response}
& 98/100 (98\%)
& 1 \texttt{substantive}$\rightarrow$\texttt{non\_substantive};
  1 \texttt{unclear}$\rightarrow$\texttt{non\_substantive}. \\

\bottomrule
\end{tabular}%
}
\end{table}
\printcredits

\bibliographystyle{cas-model2-names}

\bibliography{cas-refs-main}

@inproceedings{Guo2024DSAgent,
	title = {{DS}-{Agent}: {Automated} {Data} {Science} by {Empowering} {Large} {Language} {Models} with {Case}-{Based} {Reasoning}},
	issn = {2640-3498},
	booktitle = {Proceedings of the 41st {International} {Conference} on {Machine} {Learning}},
	publisher = {PMLR},
	author = {Guo, Siyuan and Deng, Cheng and Wen, Ying and Chen, Hechang and Chang, Yi and Wang, Jun},
	month = jul,
	year = {2024},
	pages = {16813--16848},
}

@article{Chen2026XRayScientist,
	title = {An agentic artificially intelligent {X}-ray scientist},
	volume = {8},
	issn = {2522-5839},
	doi = {10.1038/s42256-026-01261-5},
	number = {7},
	journal = {Nature Machine Intelligence},
	publisher = {Nature Publishing Group},
	author = {Chen, Zhantao and Petsch, Alexander N. and Israelski, Aidan J. and Plumley, Rajan and Shen, Lingjia and Wang, Cong and Peng, Cheng and Ni, Yuan and Bansil, Arun and Chowdhury, Sugata and Li, Mingda and Thayer, Jana B. and Thampy, Vivek and Turner, Joshua J.},
	month = jul,
	year = {2026},
	pages = {1075--1086},
}

@article{Sahu2025InsightBench,
	title = {{InsightBench}: {Evaluating} {Business} {Analytics} {Agents} {Through} {Multi}-{Step} {Insight} {Generation}},
	volume = {2025},
	language = {en},
	journal = {International Conference on Learning Representations},
	author = {Sahu, Gaurav and Puri, Abhay and Rodriguez, Juan A. and Abaskohi, Amirhossein and Chegini, Mohammad and Drouin, Alexandre and Taslakian, Perouz and Zantedeschi, Valentina and Lacoste, Alexandre and Vazquez, David and Chapados, Nicolas and Pal, Christopher and Rajeswar, Sai and Laradji, Issam},
	month = may,
	year = {2025},
	pages = {4683--4715},
}

@inproceedings{Yang2024SWEAgent,
	title = {{SWE}-agent: {Agent}-{Computer} {Interfaces} {Enable} {Automated} {Software} {Engineering}},
	volume = {37},
	doi = {10.52202/079017-1601},
	booktitle = {Advances in {Neural} {Information} {Processing} {Systems}},
	publisher = {Curran Associates, Inc.},
	author = {Yang, John and Jimenez, Carlos and Wettig, Alexander and Lieret, Kilian and Yao, Shunyu and Narasimhan, Karthik and Press, Ofir},
	year = {2024},
	pages = {50528--50652},
}

@article{Chen2025EMOS,
	title = {{EMOS}: {Embodiment}-aware {Heterogeneous} {Multi}-robot {Operating} {System} with {LLM} {Agents}},
	volume = {2025},
	journal = {International Conference on Learning Representations},
	author = {Chen, Junting and Yu, Checheng and Zhou, Xunzhe and Xu, Tianqi and Mu, Yao and Hu, Mengkang and Shao, Wenqi and Wang, Yikai and Li, Guohao and Shao, Lin},
	month = may,
	year = {2025},
	pages = {56670--56690},
}

@article{Wang2025ToolGen,
  title = {{ToolGen}: {Unified} {Tool} {Retrieval} and {Calling} via {Generation}},
  journal = {International Conference on Learning Representations},
  author = {Wang, Renxi and Han, Xudong and Ji, Lei and Wang, Shu and Baldwin, Timothy and Li, Haonan},
  year = {2025},
  pages = {73473--73498},
}

@article{Xu2025ToolLearningSurvey,
  title = {{LLM}-{Based} {Agents} for {Tool} {Learning}: {A} {Survey}},
  volume = {10},
  number = {4},
  journal = {Data Science and Engineering},
  author = {Xu, Weikai and Huang, Chengrui and Gao, Shen and Shang, Shuo},
  year = {2025},
  pages = {533--563},
  doi = {10.1007/s41019-025-00296-9},
}

@inproceedings{Yuan2025EasyTool,
  title = {{EASYTOOL}: {Enhancing} {LLM}-based {Agents} with {Concise} {Tool} {Instruction}},
  booktitle = {Proceedings of the 2025 {Conference} of the {Nations} of the {Americas} {Chapter} of the {Association} for {Computational} {Linguistics}: {Human} {Language} {Technologies} ({Volume} 1: {Long} {Papers})},
  author = {Yuan, Siyu and Song, Kaitao and Chen, Jiangjie and Tan, Xu and Shen, Yongliang and Ren, Kan and Li, Dongsheng and Yang, Deqing},
  publisher = {Association for Computational Linguistics},
  address = {Albuquerque, New Mexico},
  year = {2025},
  pages = {951--972},
  doi = {10.18653/v1/2025.naacl-long.44},
}

@inproceedings{Nanda2026Wink,
  title = {Wink: {Recovering} from {Misbehaviors} in {Coding} {Agents}},
  booktitle = {Proceedings of the 3rd {ACM} {International} {Conference} on {AI}-{Powered} {Software}},
  author = {Nanda, Rahul and Maddila, Chandra and Jha, Smriti and Khan, Euna Mehnaz and Paltenghi, Matteo and Chandra, Satish},
  publisher = {Association for Computing Machinery},
  address = {New York, NY, USA},
  year = {2026},
  pages = {208--217},
  doi = {10.1145/3805760.3814911},
}

@inproceedings{Wang2024CodeAct,
  title = {Executable {Code} {Actions} {Elicit} {Better} {LLM} {Agents}},
  booktitle = {Proceedings of the 41st {International} {Conference} on {Machine} {Learning}},
  author = {Wang, Xingyao and Chen, Yangyi and Yuan, Lifan and Zhang, Yizhe and Li, Yunzhu and Peng, Hao and Ji, Heng},
  publisher = {PMLR},
  year = {2024},
  pages = {50208--50232},
}

@misc{Yang2025AgentOccam,
  title = {{AgentOccam}: {A} {Simple} {Yet} {Strong} {Baseline} for {LLM}-{Based} {Web} {Agents}},
  author = {Yang, Ke and Liu, Yao and Chaudhary, Sapana and Fakoor, Rasool and Chaudhari, Pratik and Karypis, George and Rangwala, Huzefa},
  publisher = {arXiv},
  year = {2025},
  doi = {10.48550/arXiv.2410.13825},
  note = {arXiv:2410.13825 [cs.AI]},
}

@inproceedings{Pan2025SWEGym,
  title = {Training {Software} {Engineering} {Agents} and {Verifiers} with {SWE}-{Gym}},
  booktitle = {Proceedings of the 42nd {International} {Conference} on {Machine} {Learning}},
  author = {Pan, Jiayi and Wang, Xingyao and Neubig, Graham and Jaitly, Navdeep and Ji, Heng and Suhr, Alane and Zhang, Yizhe},
  publisher = {PMLR},
  year = {2025},
  pages = {47717--47737},
}

@misc{Jain2025R2EGym,
  title = {{R2E}-{Gym}: {Procedural} {Environments} and {Hybrid} {Verifiers} for {Scaling} {Open}-{Weights} {SWE} {Agents}},
  author = {Jain, Naman and Singh, Jaskirat and Shetty, Manish and Zheng, Liang and Sen, Koushik and Stoica, Ion},
  publisher = {arXiv},
  year = {2025},
  doi = {10.48550/arXiv.2504.07164},
  note = {arXiv:2504.07164 [cs.SE]},
}

@inproceedings{Wang2024DevilsAdvocate,
  title = {Devil's {Advocate}: {Anticipatory} {Reflection} for {LLM} {Agents}},
  booktitle = {Findings of the {Association} for {Computational} {Linguistics}: {EMNLP} 2024},
  author = {Wang, Haoyu and Li, Tao and Deng, Zhiwei and Roth, Dan and Li, Yang},
  publisher = {Association for Computational Linguistics},
  address = {Miami, Florida, USA},
  year = {2024},
  pages = {966--978},
  doi = {10.18653/v1/2024.findings-emnlp.53},
}

@inproceedings{Wang2025AgentWorkflowMemory,
  title = {Agent {Workflow} {Memory}},
  booktitle = {Proceedings of the 42nd {International} {Conference} on {Machine} {Learning}},
  author = {Wang, Zora Zhiruo and Mao, Jiayuan and Fried, Daniel and Neubig, Graham},
  publisher = {PMLR},
  year = {2025},
  pages = {63897--63911},
}

@inproceedings{Suzgun2026DynamicCheatsheet,
  title = {Dynamic {Cheatsheet}: {Test}-{Time} {Learning} with {Adaptive} {Memory}},
  booktitle = {Proceedings of the 19th {Conference} of the {European} {Chapter} of the {Association} for {Computational} {Linguistics} ({Volume} 1: {Long} {Papers})},
  author = {Suzgun, Mirac and Yuksekgonul, Mert and Bianchi, Federico and Jurafsky, Dan and Zou, James},
  publisher = {Association for Computational Linguistics},
  address = {Rabat, Morocco},
  year = {2026},
  pages = {7080--7106},
  doi = {10.18653/v1/2026.eacl-long.333},
}

@misc{Zhang2026ACE,
  title = {Agentic {Context} {Engineering}: {Evolving} {Contexts} for {Self}-{Improving} {Language} {Models}},
  author = {Zhang, Qizheng and Hu, Changran and Upasani, Shubhangi and Ma, Boyuan and Hong, Fenglu and Kamanuru, Vamsidhar and Rainton, Jay and Wu, Chen and Ji, Mengmeng and Li, Hanchen and Thakker, Urmish and Zou, James and Olukotun, Kunle},
  publisher = {arXiv},
  year = {2026},
  doi = {10.48550/arXiv.2510.04618},
  note = {arXiv:2510.04618 [cs.LG]},
}

@misc{Allard2026ERL,
  title = {Experiential {Reflective} {Learning} for {Self}-{Improving} {LLM} {Agents}},
  author = {Allard, Marc-Antoine and Teinturier, Arnaud and Xing, Victor and Viaud, Gautier},
  publisher = {arXiv},
  year = {2026},
  doi = {10.48550/arXiv.2603.24639},
  note = {arXiv:2603.24639 [cs.LG]},
}

@misc{Hu2026ContinualLearningMemory,
  title = {When {Continual} {Learning} {Moves} to {Memory}: {A} {Study} of {Experience} {Reuse} in {LLM} {Agents}},
  author = {Hu, Qisheng and Long, Quanyu and Wang, Wenya},
  publisher = {arXiv},
  year = {2026},
  doi = {10.48550/arXiv.2604.27003},
  note = {arXiv:2604.27003 [cs.LG]},
}

@misc{Bouzenia2025SEAgentTrajectories,
  title = {Understanding {Software} {Engineering} {Agents}: {A} {Study} of {Thought}-{Action}-{Result} {Trajectories}},
  author = {Bouzenia, Islem and Pradel, Michael},
  publisher = {arXiv},
  year = {2025},
  doi = {10.48550/arXiv.2506.18824},
  note = {arXiv:2506.18824 [cs.SE]},
}

@misc{Barke2026AgentRx,
  title = {{AgentRx}: {Diagnosing} {AI} {Agent} {Failures} from {Execution} {Trajectories}},
  author = {Barke, Shraddha and Goyal, Arnav and Khare, Alind and Singh, Avaljot and Nath, Suman and Bansal, Chetan},
  publisher = {arXiv},
  year = {2026},
  doi = {10.48550/arXiv.2602.02475},
  note = {arXiv:2602.02475 [cs.AI]},
}

@misc{Zhao2026PROBE,
  title = {Debugging the {Debuggers}: {Failure}-{Anchored} {Structured} {Recovery} for {Software} {Engineering} {Agents}},
  author = {Zhao, Chenyu and Zhang, Shenglin and Lin, Yihang and Gu, Wenwei and Chen, Zhimin and Sun, Yongqian and Pei, Dan and Bansal, Chetan and Rajmohan, Saravan and Ma, Minghua},
  publisher = {arXiv},
  year = {2026},
  doi = {10.48550/arXiv.2605.08717},
  note = {arXiv:2605.08717 [cs.SE]},
}

@misc{Panthi2026OutcomeMonitors,
  title = {Outcome {Monitors}: {Recovery} {Affordances} for {Silent} {Tool} {Failures}},
  author = {Panthi, Sugam and Abdelfattah, Rabab},
  publisher = {arXiv},
  year = {2026},
  doi = {10.48550/arXiv.2608.19303},
  note = {arXiv:2608.19303 [cs.AI]},
}

@misc{Dubey2026AgentFailureRepair,
  title = {Real-{Time} {Detection} and {Repair} of {LLM} {Agent} {Failures}},
  author = {Dubey, Sunny},
  publisher = {arXiv},
  year = {2026},
  doi = {10.48550/arXiv.2608.02464},
  note = {arXiv:2608.02464 [cs.AI]},
}

@article{antoniades_swe-search_2025,
	title = {{SWE}-{Search}: {Enhancing} {Software} {Agents} with {Monte} {Carlo} {Tree} {Search} and {Iterative} {Refinement}},
	volume = {2025},
	shorttitle = {{SWE}-{Search}},
	journal = {International Conference on Learning Representations},
	author = {Antoniades, Antonis and Örwall, Albert and Zhang, Kexun and Xie, Yuxi and Goyal, Anirudh and Wang, William},
	month = may,
	year = {2025},
	pages = {64485--64515},
}

@article{mu_experepair_2026,
	title = {{ExpeRepair}: {Dual}-{Memory} {Enhanced} {LLM}-{Based} {Repository}-{Level} {Program} {Repair}},
	volume = {3},
	shorttitle = {{ExpeRepair}},
	doi = {10.1145/3808181},
	number = {FSE},
	journal = {Proceedings of the ACM on Software Engineering},
	author = {Mu, Fangwen and Wang, Junjie and Shi, Lin and Wang, Song and Li, Shoubin and Wang, Qing},
	month = jun,
	year = {2026},
	pages = {FSE174:3954--FSE174:3976},
}

@inproceedings{zhang_autocoderover_2024,
	address = {New York, NY, USA},
	series = {{ISSTA} 2024},
	title = {{AutoCodeRover}: {Autonomous} {Program} {Improvement}},
	isbn = {979-8-4007-0612-7},
	shorttitle = {{AutoCodeRover}},
	doi = {10.1145/3650212.3680384},
	booktitle = {Proceedings of the 33rd {ACM} {SIGSOFT} {International} {Symposium} on {Software} {Testing} and {Analysis}},
	publisher = {Association for Computing Machinery},
	author = {Zhang, Yuntong and Ruan, Haifeng and Fan, Zhiyu and Roychoudhury, Abhik},
	month = sep,
	year = {2024},
	pages = {1592--1604},
}

@misc{yao_react_2023,
	title = {{ReAct}: {Synergizing} {Reasoning} and {Acting} in {Language} {Models}},
	shorttitle = {{ReAct}},
	doi = {10.48550/arXiv.2210.03629},
	publisher = {arXiv},
	author = {Yao, Shunyu and Zhao, Jeffrey and Yu, Dian and Du, Nan and Shafran, Izhak and Narasimhan, Karthik and Cao, Yuan},
	month = mar,
	year = {2023},
	note = {arXiv:2210.03629 [cs.CL]},
}

@misc{deepseekai2026deepseekv4,
      title={DeepSeek-V4: Towards Highly Efficient Million-Token Context Intelligence},
      author={DeepSeek-AI},
      year={2026},
}

@misc{qwen3technicalreport,
  title        = {Qwen3 Technical Report},
  author       = {{Qwen Team}},
  year         = {2025},
  eprint       = {2505.09388},
  archivePrefix= {arXiv},
  primaryClass = {cs.CL},
  url          = {https://arxiv.org/abs/2505.09388}
}

@misc{sonwane_omnicode_2026,
	title = {{OmniCode}: {A} {Benchmark} for {Evaluating} {Software} {Engineering} {Agents}},
	shorttitle = {{OmniCode}},
	doi = {10.48550/arXiv.2602.02262},
	publisher = {arXiv},
	author = {Sonwane, Atharv and Tu, Eng-Shen and Lu, Wei-Chung and Beger, Claas and Larsen, Carter and Dhar, Debjit and Alford, Simon and Chen, Rachel and Pattanayak, Ronit and Dang, Tuan Anh and Chen, Guohao and Geng, Gloria and Ellis, Kevin and Dutta, Saikat},
	month = may,
	year = {2026},
	note = {arXiv:2602.02262 [cs.SE]},
}

@misc{nevill-manning_identifying_1997,
	title = {Identifying {Hierarchical} {Structure} in {Sequences}: {A} linear-time algorithm},
	shorttitle = {Identifying {Hierarchical} {Structure} in {Sequences}},
	doi = {10.48550/arXiv.cs/9709102},
	publisher = {arXiv},
	author = {Nevill-Manning, C. G. and Witten, I. H.},
	month = sep,
	year = {1997},
	note = {arXiv:cs/9709102},
}

@inproceedings{yeh_matrix_2016,
	title = {Matrix {Profile} {I}: {All} {Pairs} {Similarity} {Joins} for {Time} {Series}: {A} {Unifying} {View} {That} {Includes} {Motifs}, {Discords} and {Shapelets}},
	issn = {2374-8486},
	shorttitle = {Matrix {Profile} {I}},
	doi = {10.1109/ICDM.2016.0179},
	booktitle = {2016 {IEEE} 16th {International} {Conference} on {Data} {Mining} ({ICDM})},
	author = {Yeh, Chin-Chia Michael and Zhu, Yan and Ulanova, Liudmila and Begum, Nurjahan and Ding, Yifei and Dau, Hoang Anh and Silva, Diego Furtado and Mueen, Abdullah and Keogh, Eamonn},
	month = dec,
	year = {2016},
	note = {ISSN: 2374-8486},
	pages = {1317--1322},
}

@misc{mehtiyev_beyond_2026,
	title = {Beyond {Resolution} {Rates}: {Behavioral} {Drivers} of {Coding} {Agent} {Success} and {Failure}},
	shorttitle = {Beyond {Resolution} {Rates}},
	doi = {10.48550/arXiv.2604.02547},
	publisher = {arXiv},
	author = {Mehtiyev, Tural and Assunção, Wesley},
	month = apr,
	year = {2026},
	note = {arXiv:2604.02547 [cs.SE]},
}

@inproceedings{chen-etal-2024-beyond-natural,
    title = "Beyond Natural Language: {LLM}s Leveraging Alternative Formats for Enhanced Reasoning and Communication",
    author = "Chen, Weize  and
      Yuan, Chenfei  and
      Yuan, Jiarui  and
      Su, Yusheng  and
      Qian, Chen  and
      Yang, Cheng  and
      Xie, Ruobing  and
      Liu, Zhiyuan  and
      Sun, Maosong",
    editor = "Al-Onaizan, Yaser  and
      Bansal, Mohit  and
      Chen, Yun-Nung",
    booktitle = "Findings of the Association for Computational Linguistics: EMNLP 2024",
    month = nov,
    year = "2024",
    address = "Miami, Florida, USA",
    publisher = "Association for Computational Linguistics",
    doi = "10.18653/v1/2024.findings-emnlp.623",
    pages = "10626--10641",
}

@misc{qwen3.6-27b,
    title  = {{Qwen3.6-27B}: Flagship-Level Coding in a {27B} Dense Model},
    author = {{Qwen Team}},
    month  = {April},
    year   = {2026},
    url    = {https://qwen.ai/blog?id=qwen3.6-27b}
}

@misc{qwen37,
    title = {{Qwen3.7}: The Agent Frontier},
    url = {https://qwen.ai/blog?id=qwen3.7},
    author = {{Qwen Team}},
    month = {May},
    year = {2026}
}

@misc{kimiteam2026kimik3openfrontier,
      title={Kimi K3: Open Frontier Intelligence}, 
      author={Kimi Team and Tongtong Bai and Yifan Bai and Yiping Bao and others},
      year={2026},
      eprint={2607.24653},
      archivePrefix={arXiv},
      primaryClass={cs.CL},
}

@misc{sinha2024realtimeanomalydetectionreactive,
      title={Real-Time Anomaly Detection and Reactive Planning with Large Language Models}, 
      author={Rohan Sinha and Amine Elhafsi and Christopher Agia and Matthew Foutter and Edward Schmerling and Marco Pavone},
      year={2024},
      eprint={2407.08735},
      archivePrefix={arXiv},
      primaryClass={cs.RO},
}

@inproceedings{li-etal-2026-towards,
    title = "Towards Self-Improving Error Diagnosis in Multi-Agent Systems",
    author = "Li, Jiazheng  and
      Yilmaz, Emine  and
      Chen, Bei  and
      Le, Thu",
    editor = "Liakata, Maria  and
      Moreira, Viviane P.  and
      Zhang, Jiajun  and
      Jurgens, David",
    booktitle = "Findings of the {A}ssociation for {C}omputational {L}inguistics: {ACL} 2026",
    month = jul,
    year = "2026",
    address = "San Diego, California, United States",
    publisher = "Association for Computational Linguistics",
    doi = "10.18653/v1/2026.findings-acl.98",
    pages = "2063--2077",
    ISBN = "979-8-89176-395-1",
}

@misc{liu2026onlinemonitoringcorrectivesteering,
      title={Online Monitoring and Corrective Steering of Programming Agents}, 
      author={Shuyang Liu and Saman Dehghan and Ji Young Kim and Jatin Ganhotra and Martin Hirzel and Reyhaneh Jabbarvand},
      year={2026},
      eprint={2608.06701},
      archivePrefix={arXiv},
      primaryClass={cs.SE},
}

@book{dewey1916democracy,
  author    = {Dewey, John},
  title     = {Democracy and Education:
               An Introduction to the Philosophy of Education},
  year      = {1916},
  publisher = {The Macmillan Company},
  address   = {New York}
}

@book{kolb1984experiential,
  author    = {Kolb, David A.},
  title     = {Experiential Learning: Experience as the Source
               of Learning and Development},
  year      = {1984},
  publisher = {Prentice-Hall},
  address   = {Englewood Cliffs, NJ},
  isbn      = {9780132952613}
}

@inproceedings{
    jimenez2024swebench,
    title={{SWE}-bench: Can Language Models Resolve Real-world Github Issues?},
    author={Carlos E Jimenez and John Yang and Alexander Wettig and Shunyu Yao and Kexin Pei and Ofir Press and Karthik R Narasimhan},
    booktitle={The Twelfth International Conference on Learning Representations},
    year={2024},
}

@misc{liu2025agentbenchevaluatingllmsagents,
      title={AgentBench: Evaluating LLMs as Agents}, 
      author={Xiao Liu and Hao Yu and Hanchen Zhang and Yifan Xu and Xuanyu Lei and Hanyu Lai and Yu Gu and Hangliang Ding and Kaiwen Men and Kejuan Yang and Shudan Zhang and Xiang Deng and Aohan Zeng and Zhengxiao Du and Chenhui Zhang and Sheng Shen and Tianjun Zhang and Yu Su and Huan Sun and Minlie Huang and Yuxiao Dong and Jie Tang},
      year={2025},
      eprint={2308.03688},
      archivePrefix={arXiv},
      primaryClass={cs.AI},
}

@misc{Young2025Harness,
  author       = {Young, Justin},
  title        = {Effective Harnesses for Long-Running Agents},
  year         = {2025},
  month        = nov,
  howpublished = {Anthropic Engineering},
  note         = {Published November 26, 2025}
}

@misc{Ji2026Sentry,
  author        = {Ji, Changxiu and Lu, Amy and Zhang, Qizheng and Olukotun, Kunle},
  title         = {{Sentry}: Learning to Recover from {LLM} Agent Failures at Test Time},
  year          = {2026},
  eprint        = {2610.02994},
  archivePrefix = {arXiv},
  primaryClass  = {cs.LG},
  doi           = {10.48550/arXiv.2610.02994}
}

@misc{wang2026openhandssoftwareagentsdk,
      title={The OpenHands Software Agent SDK: A Composable and Extensible Foundation for Production Agents}, 
      author={Xingyao Wang and Simon Rosenberg and Juan Michelini and Calvin Smith and Hoang Tran and Engel Nyst and Rohit Malhotra and Xuhui Zhou and Valerie Chen and Robert Brennan and Graham Neubig},
      year={2026},
      eprint={2511.03690},
      archivePrefix={arXiv},
      primaryClass={cs.SE},
}

@misc{duan2026interveneempiricalstudyerrors,
      title={When to Intervene? An Empirical Study of Errors and Self-Recovery in LLM Agents}, 
      author={Chiming Duan and Tong Jia and Yingxi Chen and Minghua He and Lingzhe Zhang and Ying Li},
      year={2026},
      eprint={2610.36471},
      archivePrefix={arXiv},
      primaryClass={cs.AI},
}

@inproceedings{li-etal-2023-api,
    title = "{API}-Bank: A Comprehensive Benchmark for Tool-Augmented {LLM}s",
    author = "Li, Minghao  and
      Zhao, Yingxiu  and
      Yu, Bowen  and
      Song, Feifan  and
      Li, Hangyu  and
      Yu, Haiyang  and
      Li, Zhoujun  and
      Huang, Fei  and
      Li, Yongbin",
    booktitle = "Proceedings of the 2023 Conference on Empirical Methods in Natural Language Processing",
    month = dec,
    year = "2023",
    address = "Singapore",
    publisher = "Association for Computational Linguistics",
    doi = "10.18653/v1/2023.emnlp-main.187",
    pages = "3102--3116"
}

@inproceedings{pmlr-v267-patil25a,
  title = 	 {The Berkeley Function Calling Leaderboard ({BFCL}): From Tool Use to Agentic Evaluation of Large Language Models},
  author =       {Patil, Shishir G and Mao, Huanzhi and Yan, Fanjia and Ji, Charlie Cheng-Jie and Suresh, Vishnu and Stoica, Ion and Gonzalez, Joseph E.},
  booktitle = 	 {Proceedings of the 42nd International Conference on Machine Learning},
  pages = 	 {48371--48392},
  year = 	 {2025},
  volume = 	 {267},
  series = 	 {Proceedings of Machine Learning Research},
  month = 	 {13--19 Jul},
  publisher =    {PMLR}
}

@misc{yao2024taubenchbenchmarktoolagentuserinteraction,
      title={$\tau$-bench: A Benchmark for Tool-Agent-User Interaction in Real-World Domains}, 
      author={Shunyu Yao and Noah Shinn and Pedram Razavi and Karthik Narasimhan},
      year={2024},
      eprint={2406.12045},
      archivePrefix={arXiv},
      primaryClass={cs.AI},
      url={https://arxiv.org/abs/2406.12045}, 
}

\end{document}